\documentclass[11pt]{article}

\usepackage[final]{acl}
\usepackage{times}
\usepackage{latexsym}
\usepackage[T1]{fontenc}
\usepackage[utf8]{inputenc}
\usepackage{microtype}
\usepackage{inconsolata}
\usepackage{graphicx}
\usepackage{amsmath}
\usepackage{amssymb}
\usepackage{booktabs}
\usepackage{listings}
\usepackage{xcolor}
\usepackage{pifont}
\usepackage{makecell}
\usepackage{multirow}

\usepackage{colortbl}
\usepackage{array}
\definecolor{goldcol}{gray}{0.96}
\definecolor{deltacol}{RGB}{235,245,255}
\newcolumntype{G}{>{\columncolor{goldcol}}r}
\newcolumntype{D}{>{\columncolor{deltacol}}r}

\title{Diagnostic Analysis for Large Language Model Reasoning}
\author{
  Xinyu Liu\textsuperscript{1} \quad
  Dongqi Fu\textsuperscript{2} \quad
  Jun Wu\textsuperscript{1} \\
  \textsuperscript{1}Michigan State University \quad
  \textsuperscript{2}Meta AI \\
  \texttt{\{liuxin73, wujun4\}@msu.edu} \quad
  \texttt{dongqifu@meta.com}
}

\begin{document}
\maketitle

\begin{abstract}
Reasoning structures (e.g., chains and graphs) are increasingly used for probabilistic inference, factuality verification, and uncertainty estimation, but their structural reliability is rarely evaluated directly.
We test whether this reliability holds through a proposed two-stage diagnostic evaluation
in open-ended multi-hop QA. Stage~1 tests different capabilities required to construct reasoning graphs, showing that graph reliability varies with topology, prompt design, and model family. 
Stage~2 uses uncertainty estimation as a downstream test case. Generated-edge experiments show that graph-based computation is sensitive to upstream graph errors, with performance dropping relative to the gold-graph setting. Gold-graph experiments further show that graph structure provides little benefit when used only as prompt context, but improves the separation of correct from incorrect answers by confidence score and enables node-level diagnostics.

% \dq{also, i may suggest we name stage 1 and stage 2, instead of solely using stage 1 and stage 2 in abs and intro}

\end{abstract}

\section{Introduction}
\label{sec:introduction}
% \vspace{-0.15cm}
Recent work~\cite{xiong-etal-2025-mapping,besta2024graph,feng_bird_2025,da_understanding_2025} increasingly represents LLM reasoning as graph-structured intermediate states, where nodes represent intermediate claims, sub-questions, latent variables, or reasoning steps, while edges encode dependency, deducibility, or probabilistic influence. The resulting graph is then used as a scaffold for downstream procedures such as probabilistic inference and calibrated prediction \citep{huang_bayesagent_2026, zhang_bayesian_nodate, feng_bird_2025, nafar_extracting_2025}, factuality verification \citep{rubinconformal}, and uncertainty estimation \citep{da_understanding_2025}. This line of work therefore relies on a shared assumption: the prompted graph is reliable enough for the computation built on top of it.

That assumption is rarely tested directly. A generated reasoning graph can fail at several levels: it may omit necessary steps, connect the wrong dependencies, introduce shortcut edges, assign incorrect intermediate answers, or attach unsupported evidence. These errors are not cosmetic. A wrong dependency can change a probabilistic factorization; an incorrect node answer can contaminate downstream reasoning; missing evidence can make a graph appear grounded when it is not. Yet many uses of graph-structured reasoning evaluate the downstream output more than the graph construction process itself, and existing graph-level checks are often limited in scale or do not isolate graph-specific failure modes \citep{rubinconformal, zhang_bayesian_nodate}.

To evaluate this assumption, we need to separate capabilities that are usually bundled together in graph generation. To be specific, it involves decomposing a question into reasoning steps, connecting them with correct logical dependencies, generating sub-answers at each node, and grounding each step in supporting evidence. Failures along these axes have different downstream implications: miswired edges can corrupt probabilistic inference, while evidence errors undermine factual grounding. To isolate them, we represent each graph node as a (sub-question, sub-answer, evidence) triple and evaluate models under three controlled settings on multi-hop QA datasets with gold scaffolding.

We study this problem in two stages. Stage~1 evaluates whether LLMs can construct reliable reasoning graphs across the three diagnostic settings. Stage~2 uses uncertainty estimation as a concrete downstream setting where graph quality directly impacts performance and asks how graph structure affects final-answer confidence. In Stage~2, we ask whether graph structure helps more as prompt context or as an explicit computation structure, and test how construction errors propagate to confidence quality.

% \dq{the following paragraph is vague and not very clear}

Our results give a qualified answer. Frontier models can construct high-quality reasoning graphs under sufficient prompting, but graph reliability is not a stable property: smaller models degrade substantially, and performance is sensitive to model family, graph topology, and prompt design. For uncertainty estimation, providing the graph as prompt context gives limited benefit, and verbalized confidence is often poorly discriminative. Using the graph as an explicit computation structure provides useful confidence and enables node-level diagnostics, but its advantage depends on graph quality: when gold edges are replaced with generated edges, graph errors directly reduce downstream performance.

Our contributions are summarized as follows\footnote{
Code is publicly available at
\url{https://github.com/hint-group/llm-reasoning-graph-diagnostics.git}.
}:

\begin{itemize}
    \item \textbf{A staged evaluation for reasoning graph reliability.}
    We introduce a controlled evaluation that tests different capabilities required to construct reasoning graphs, allowing prompted graphs to be evaluated as intermediate structures rather than only through downstream task performance.

    \item \textbf{A structured account of graph-generation failures.}
    We show that graph reliability varies across models, graph structures, and prompt schema rather than following model size alone. More importantly, when the task moves beyond connecting given steps to generating node content and full decompositions, failures shift from dependency errors to grounding and decomposition errors.

    \item \textbf{A controlled comparison of graph-use strategies for uncertainty estimation.}
    We provide a systematic comparison of graph-free, graph-as-prompt, verbalized-confidence, and graph-as-computation methods under gold graphs, and add a generated-edge stress test that measures how construction errors propagate to downstream confidence quality.

\end{itemize}
% \vspace{0.1cm}
\section{Related Work}
\label{sec:related}

\paragraph{Graph-structured LLM reasoning.}
Recent work increasingly models LLM reasoning as graph-structured objects that can be constructed by prompting and used for downstream inference or analysis.
\citet{rubinconformal} construct deducibility graphs over atomic claims to support conformal factuality guarantees, with graph validation performed on a small set of examples.
\citet{da_understanding_2025} elicit reasoning-topology graphs to study uncertainty in LLM explanations, but the reliability of the elicited topology is not the main object of evaluation.
Complementary benchmark work also shows that reasoning structure matters: \citet{pahilajani_grs-qa_2024} introduce graph-annotated QA examples and find that LLM performance varies across reasoning structures.
Together, these works motivate graph-level evaluation, but they leave open a different question: when the graph is itself generated or used as a computational scaffold, how reliable are its nodes, edges, evidence links, and decompositions?

\paragraph{LLMs and probabilistic graphical models.}
A related line of work uses LLMs to construct, parameterize, or reason over probabilistic graphical models. 
\citet{huang_bayesagent_2026} elicit latent uncertainty variables and dependencies to form verbalized probabilistic graphical models for calibrated prediction. 
\citet{feng_bird_2025} align Bayesian networks with LLM-based abductive reasoning for probability estimation in decision-making. 
\citet{zhang_bayesian_nodate} use LLM-guided search to induce Bayesian network structures and evaluate them against domain knowledge. 
\citet{nafar_extracting_2025} query LLMs for conditional probability table entries to parameterize Bayesian networks. 
These approaches make graphical structure part of the probabilistic computation, so errors in generated structures or verbalized probabilities can affect downstream estimates. 
We therefore treat verbalized probability elicitation and graph-based computation as empirical alternatives rather than assuming either is reliable.

\paragraph{Uncertainty estimation for LLM reasoning.}
LLM uncertainty estimation is often built from either sampling behavior or verbalized confidence. 
Self-consistency \citep{wang2022self} aggregates multiple reasoning samples; the resulting answer frequency is commonly used as a sampling-based confidence signal. 
Verbalized approaches instead ask models to report confidence directly \citep{lin2022teaching,tian2023just}, but such estimates can be poorly calibrated and sensitive to prompting; \citet{tanneru_quantifying_2023} show that verbalized estimates may be unreliable proxies for explanation confidence. 
Recent surveys and benchmarks further emphasize that LLM uncertainty includes reasoning-path variability and decoding stochasticity, not only final-answer probability \citep{liu2025uncertainty,vashurin2025benchmarking}. 
This motivates comparing confidence signals that come from sampling, verbalization, and graph-based computation rather than treating uncertainty as a single scalar elicitation problem.
\section{Diagnostic Analysis for LLM Reasoning}
\label{sec:method}

% \wu{always start the main section from the problem studied in this work. In this paper, we focus on diagnosing LLM reasoning.
% A good roadmap for this section can be: LLM reasoning -> reasoning structure analysis -> two-stage diagnostic framework.
% }

% \wu{For example, section 3.1: summarize problem of LLM reasoning, section 3.2: explain why DAG structure is involved (these two subsections also can be combined), section 3.3: give an overview of the proposed two-stage diagnostic framework: stage 1: diagnosing reasoning structure, stage 2: diagnosing reasoning uncertainty/confidence}

% \wu{The settings A-C can move to section 4, as those three settings are mainly used for reasoning structure analysis}

\subsection{Reasoning Graphs as Diagnostic Objects}
\label{sec:graph-representation}

We diagnose the reasoning graph as an output artifact, not the model's hidden reasoning process. The graph may be produced jointly with an answer or constructed post-hoc from an LLM reasoning trace. In either case, our evaluation asks whether the resulting graph satisfies its stated semantics: nodes should represent meaningful reasoning steps with correct local answers and evidence, and edges should represent immediate task dependencies. 
% This scope separates graph reliability from task success or explanation plausibility.

We represent a reasoning graph as a directed acyclic graph $G=(V,E)$. Each node $v_i \in V$ denotes a reasoning step, represented by a sub-question $q_i$, an intermediate answer $a_i$, and a set of evidence identifiers $\mathcal{E}_i$ that index supporting paragraphs in the input context. Each directed edge $e_{ij} \in E$ indicates that answering step $j$ requires the answer from step $i$. The graph includes a distinguished final node $v_{\mathrm{final}}$ whose question $q_{\mathrm{final}}$ is the original question and whose answer $a_{\mathrm{final}}$ is the target output.

The DAG structure encodes data-flow dependency, not only generation order. Because answering $v_j$ requires $a_i$ as input, errors at $v_i$ propagate forward to all downstream nodes. Acyclicity guarantees a valid topological ordering and enables a factored probabilistic model over node answers. Merge nodes where two or more parent chains converge are structurally consequential: a single miswired edge at such a junction can corrupt the factorization and contaminate the final answer. We therefore treat edges as immediate dependencies: if $v_1$ supports $v_2$ and $v_2$ supports $v_{\mathrm{final}}$, the direct edge $v_1 \to v_{\mathrm{final}}$ is a shortcut unless the final answer depends on $v_1$ directly.

\subsection{Two-Stage Diagnostic Protocol}
\label{sec:framework}

Our diagnostic protocol separates two questions that are often conflated. First, can an LLM produce a reasoning graph whose nodes, evidence, and dependency edges satisfy the intended graph semantics? Second, if such a graph is available, how does its structure affect downstream uncertainty estimation? We address the first question with Stage~1, which evaluates graph construction under progressively weaker gold scaffolding. We address the second with Stage~2, which uses uncertainty estimation as a controlled downstream test case under both gold and generated graph structures.

\textbf{Stage~1} organizes evaluation into three controlled settings that progressively remove gold scaffolding (Table~\ref{tab:diagnostic-settings}).

% \vspace{-0.15cm}
\begin{table}[htbp]
\centering
\footnotesize
\resizebox{\linewidth}{!}{
\begin{tabular}{lp{2.2cm}p{1.8cm}p{2.0cm}}
\toprule
\textbf{Set.} & \textbf{Given to model} & \textbf{Predicted} & \textbf{Purpose} \\
\midrule
A & Gold sub-questions, answers, paragraphs & Edges & Dependency wiring \\
B & Gold sub-questions, paragraphs & Answers, evidence IDs, edges & Local node grounding \\
C & Question, paragraphs & Full graph & End-to-end generation \\
\bottomrule
\end{tabular}}
\caption{Three controlled diagnostic settings. Difficulty increases from A to C as gold scaffolding is progressively removed.}
\label{tab:diagnostic-settings}
\end{table}
% \vspace{-0.15cm}

Setting~A fixes local node states and isolates dependency wiring. Setting~B keeps the gold decomposition but requires the model to generate node answers, evidence, and edges. Setting~C removes the decomposition scaffold entirely. LLM needs to generate the full graph from the original question and candidate paragraphs. The A$\rightarrow$B gap measures the cost of local node grounding under a fixed decomposition; the B$\rightarrow$C gap measures the additional cost of generating the decomposition itself.

\textbf{Stage~2} uses uncertainty estimation as a downstream test case for graph reliability. We first fix the gold graph to isolate graph-use strategy from graph-generation errors, comparing four approaches: graph-free sampling, graph-as-prompt, verbalized confidence, and graph-as-computation. We then replace gold edges with Stage~1 generated edges to measure how upstream construction errors propagate to confidence quality, and stratify results by hop count and topology to identify the structural conditions under which each strategy is effective.

\section{Stage~1: Reasoning Graph Evaluation}
\label{sec:experiments}

Stage~1 applies the three diagnostic settings (Table~\ref{tab:diagnostic-settings}) to two multi-hop QA benchmarks to characterize when and how LLMs induce reliable reasoning graphs. The key question is whether graph-generation capability is a stable, scalable property or whether it dissociates across models, settings, and structures.

% \vspace{-0.1cm}
% ============================================================
\subsection{Experimental Setup}
\label{sec:stage1-setup}
% ============================================================
\paragraph{Datasets.}
Stage~1 uses two multi-hop QA benchmarks. MuSiQue \citep{trivedi2022musique} provides gold sub-question decompositions, dependency structures, and paragraph-level evidence annotations. We evaluate over the full development set (2,417 examples) with different hop counts and structures. We also evaluate a stratified 2WikiMultiHopQA subset \citep{ho-etal-2020-constructing} across all three diagnostic settings (498 examples). Because 2WikiMultiHopQA does not supply explicit reasoning nodes like MuSiQue does, we construct gold nodes from evidence triples and map them to context-paragraph indices as evidence identifiers. The subset is stratified across four question types: comparison, bridge\_comparison, compositional, and inference. 
% Full results on 2WikiMultiHopQA  appear in Appendix~\ref{app:twowiki-stage1}.

% \vspace{-0.15cm}
\paragraph{Models.}
We evaluate nine models spanning three open-source families and two frontier closed models: Llama-3.1-8B-Instruct and Llama-3.1-70B-Instruct \citep{grattafiori2024llama}; Qwen3-8B, Qwen3-14B, and Qwen3-32B in non-thinking mode, plus Qwen3-32B-Reasoning with thinking mode enabled \citep{yang2025qwen3}; DeepSeek V4-Flash \citep{deepseekai2026deepseekv4}; Claude Sonnet~4.5 \citep{anthropic2025claude45system}; and GPT-5 \citep{singh2025openai}.

% \paragraph{Reproducibility.}
% We use temperature~0 for all Stage~1 evaluations. Full prompt text and output schemas are in Appendix~\ref{app:prompts}. Prompt SHA256 checksums are logged per run. Parse rates and schema validity are reported in Appendix~\ref{app:reproducibility}.\wu{This paragraph can be moved to appendix if space is limited.}

% \paragraph{Metrics.}
% In Settings~A and B, the node set is fixed, so dependency wiring is evaluated by edge precision, recall, F1, and exact edge-set match. Setting~B additionally evaluates generated node content using Ans~F1 over sub-answers and Attr~F1 over supporting paragraph IDs. In Setting~C, models generate their own decompositions, so predicted nodes are aligned to gold nodes using semantic cosine similarity. We report HPS~(Hit Per Step) (\citealt{ning_mc-search_2026}), the fraction of gold reasoning steps recovered; Edge HPS, a reachability-based recall metric over gold dependencies; Step~F1, token overlap between aligned predicted and gold sub-questions; and $\Delta$Step, the difference between predicted and gold node counts. Full alignment and metric details are in Appendix~\ref{app:metrics}.

% \vspace{-0.15cm}
\paragraph{Metrics.}
For Settings~A and B, the node set is fixed, so dependency wiring is evaluated over directed edge sets. Given gold edges $E$ and predicted edges $\hat{E}$, we report edge F1 (set F1 between $\hat{E}$ and $E$) and exact edge match $\mathbb{I}[\hat{E}=E]$, where $\mathbb{I}[\cdot]$ is the identity function. For Setting~B, Ans~F1 macro-averages token-overlap F1 between predicted and gold sub-answers over matched nodes, while Attr~F1 is micro-averaged within each example over all matched non-final nodes, then macro-averaged across examples.
In Setting~C, let $V$ and $\hat{V}$ denote the gold and predicted node sets. Predicted nodes are aligned to gold nodes by semantic matching using
e5-large-v2 embeddings with cosine-similarity threshold $\tau=0.85$;
let $M \subseteq V \times \hat{V}$ be the matched pairs. HPS~(Hit Per Step; \citealt{ning_mc-search_2026}) is $|M|/|V|$, the fraction of gold reasoning steps recovered. Step~F1 is step-level F1 between $\hat{V}$ and $V$ under semantic matching. Edge HPS is recall over gold dependency edges after node alignment: a gold edge is recovered when both endpoints are matched and a directed path exists between the corresponding predicted nodes. Finally, $\Delta\mathrm{Step}=|\hat{V}|-|V|$ measures over- or under-decomposition. Sensitivity to the alignment threshold and embedding backbone is analyzed
in Appendix~\ref{app:alignment-sensitivity}. Full details are in Appendix~\ref{app:metrics}.

% Table 1: Stage 1 cross-model performance under v2 prompts, MuSiQue full dev (n=2417).
% A = dependency wiring with gold nodes;
% B = fixed sub-questions with answer/evidence/edge generation;
% C = free decomposition.
\begin{table*}[t]
\centering
\scriptsize
\setlength{\tabcolsep}{4pt}
\resizebox{\textwidth}{!}{%
\begin{tabular}{llr rr rrr rrrr}
\toprule
& & &
  \multicolumn{2}{c}{\textbf{Setting A}} &
  \multicolumn{3}{c}{\textbf{Setting B}} &
  \multicolumn{4}{c}{\textbf{Setting C}} \\
\cmidrule(lr){4-5}\cmidrule(lr){6-8}\cmidrule(lr){9-12}
\textbf{Model} & \textbf{Family} & \textbf{Params} &
  \textbf{Edge F1} & \textbf{Exact Edge} &
  \textbf{Edge F1} & \textbf{Ans F1} & \textbf{Attr F1} &
  \textbf{HPS} & \textbf{Step F1} & \textbf{Edge HPS} & \textbf{Attr F1} \\
\midrule
Llama-3.1-8B  & Llama  & 8B  &
  0.797$^\dagger$ & 0.388$^\dagger$ &
  0.803 & 0.386 & 0.567 &
  0.688 & 0.679 & 0.344 & 0.676 \\
Qwen3-8B      & Qwen   & 8B  &
  0.809$^\dagger$ & 0.444$^\dagger$ &
  0.970 & 0.504 & 0.706 &
  0.735 & 0.777 & 0.533 & 0.786 \\
Qwen3-14B     & Qwen   & 14B &
  0.963 & 0.913 &
  0.957 & 0.465 & 0.763 &
  0.801 & 0.837 & 0.567 & 0.812 \\
Qwen3-32B     & Qwen   & 32B &
  0.985 & 0.961 &
  0.970 & 0.419 & 0.795 &
  0.863 & 0.857 & 0.705 & 0.831 \\
Qwen3-32B-Reasoning & Qwen & 32B &
  0.992 & 0.976 &
  0.973 & 0.579 & 0.821 &
  0.878 & 0.894 & 0.745 & 0.850 \\
Llama-3.1-70B & Llama  & 70B &
  0.899 & 0.778 &
  0.909 & 0.625 & 0.813 &
  0.800 & 0.834 & 0.612 & 0.829 \\
DeepSeek V4-Flash & DeepSeek & ---$^\ddagger$ &
  0.992 & 0.981 &
  0.985 & 0.674 & 0.839 &
  0.861 & 0.869 & 0.692 & 0.861 \\
GPT-5 & GPT & --- &
  0.986 & 0.967 &
  0.981 & 0.468 & 0.853 &
  0.934 & 0.903 & 0.809 & 0.882 \\
Claude Sonnet 4.5 & Claude & --- &
  \textbf{0.994} & \textbf{0.986} &
  \textbf{0.991} & \textbf{0.697} & \textbf{0.882} &
  \textbf{0.960} & \textbf{0.933} & \textbf{0.890} & \textbf{0.895} \\
% \midrule

\bottomrule
\end{tabular}}
\caption{
Stage~1 results on MuSiQue full dev.
Bold marks the best completed value in each column.
$^\dagger$ Subject to schema-example artifact in the edge-only setting; see Appendix~\ref{app:schema-sensitivity}.
$^\ddagger$ DeepSeek V4-Flash is a mixture-of-experts model with 284B total parameters and 13B activated parameters.
}
\label{tab:stage1-main-results}
\end{table*}
 
\begin{table*}[t]
\centering
\small
\resizebox{\textwidth}{!}{
\begin{tabular}{lrrrrrrrrrr}
\toprule
\textbf{Slice} & \textbf{N}
& \multicolumn{3}{c}{\textbf{Llama-3.1-70B}}
& \multicolumn{3}{c}{\textbf{Qwen3-32B}}
& \multicolumn{3}{c}{\textbf{Claude Sonnet 4.5}} \\
\cmidrule(lr){3-5}
\cmidrule(lr){6-8}
\cmidrule(lr){9-11}
& & \textbf{HPS} & \textbf{Edge HPS} & $\boldsymbol{\Delta}$ \textbf{Step}
& \textbf{HPS} & \textbf{Edge HPS} & $\boldsymbol{\Delta}$ \textbf{Step}
& \textbf{HPS} & \textbf{Edge HPS} & $\boldsymbol{\Delta}$ \textbf{Step} \\
\midrule
2-hop & 1252 & 0.918 & 0.780 & 0.049 & 0.921 & 0.820 & 0.133 & 0.973 & 0.920 & 0.159 \\
3-hop & 760 & 0.716 & 0.480 & -0.622 & 0.842 & 0.638 & -0.041 & 0.962 & 0.882 & 0.217 \\
4-hop & 405 & 0.590 & 0.340 & -1.062 & 0.721 & 0.478 & -0.397 & 0.917 & 0.814 & 0.165 \\\midrule
simple\_tree & 1820 & 0.855 & 0.698 & -0.162 & 0.897 & 0.780 & 0.069 & 0.968 & 0.910 & 0.169 \\
v\_structure & 351 & 0.683 & 0.350 & -0.723 & 0.792 & 0.454 & -0.108 & 0.957 & 0.836 & 0.393 \\
complex\_tree & 246 & 0.556 & 0.349 & -1.193 & 0.714 & 0.511 & -0.459 & 0.907 & 0.821 & -0.061 \\
\bottomrule
\end{tabular}}
\caption{Setting~C end-to-end graph generation performance stratified by hop count and topology slices.}
\label{tab:topology-strat}
\end{table*}
% ============================================================
\subsection{Capability Dissociation Across Models and Settings}
\label{sec:stage1-capability}
% ============================================================

Table~\ref{tab:stage1-main-results} reports MuSiQue results across nine models and all three settings; we examine how the three capability axes relate within and across different models.

\paragraph{Within-family scale dissociates edge wiring from answer/evidence grounding.}

Within Qwen3, Setting~A Edge F1 scales monotonically with parameter count, from 0.809 at 8B to 0.985 at 32B.\footnote{The Qwen3-8B Edge F1 under the E1 schema is confounded by a 47.2\% disconnected-graph rate from schema-example mimicry; under E3 it recovers to approximately 0.92. See Appendix~\ref{app:schema-sensitivity}.} Setting~B Ans F1 does not follow: 8B achieves 0.504, but 14B drops to 0.465 and 32B drops further to 0.419. The two metrics move in opposite directions as scale increases, indicating that dependency wiring and answer generation do not co-scale within a single model family. Enabling reasoning mode partially rescues answer grounding: Qwen3-32B-Reasoning reaches B Ans F1 0.579, reducing but not closing the gap to frontier performance. A Setting~A comparison of thinking-on versus non-thinking mode appears in Appendix~\ref{app:thinking}.

\paragraph{Cross-family profiles confirm the dissociation.}

The most direct cross-family contrast is Qwen3-32B versus Llama-3.1-70B in Setting~B: Qwen3-32B achieves Edge F1 0.970 but Ans F1 0.419, while Llama-3.1-70B achieves Edge F1 0.909 but Ans F1 0.625. The two models rank in opposite order depending on which axis is measured. GPT-5 instantiates the same pattern: Setting~A Edge F1 0.986 but Setting~B Ans F1 only 0.468, below Llama-3.1-70B despite stronger structural performance. In Setting~B, strong edge wiring does not guarantee strong node-answer or evidence grounding. Notably, Llama-3.1-70B trails Qwen3-32B by 8.6 Edge F1 points in Setting~A despite having twice the parameters, confirming that the dissociation reflects model-family effects rather than scale alone.

\paragraph{Dissociation on 2WikiMultiHopQA.}

The same pattern replicates on 2WikiMultiHopQA: Qwen3-32B leads on structural metrics while Llama-3.1-70B leads on answer generation; full results appear in Appendix~\ref{app:twowiki-stage1}.

% ============================================================
\subsection{Structural Difficulty Across Datasets}
\label{sec:stage1-structural-difficulty}
% ============================================================
% \begin{table*}[t]
% \centering
% \small
% \resizebox{\textwidth}{!}{
% \begin{tabular}{lrrrrrrrrrr}
% \toprule
% \textbf{Slice} & \textbf{N}
% & \multicolumn{3}{c}{\textbf{Llama-3.1-70B}}
% & \multicolumn{3}{c}{\textbf{Qwen3-32B}}
% & \multicolumn{3}{c}{\textbf{Claude Sonnet 4.5}} \\
% \cmidrule(lr){3-5}
% \cmidrule(lr){6-8}
% \cmidrule(lr){9-11}
% & & \textbf{HPS} & \textbf{Edge HPS} & $\boldsymbol{\Delta}$ \textbf{Step}
% & \textbf{HPS} & \textbf{Edge HPS} & $\boldsymbol{\Delta}$ \textbf{Step}
% & \textbf{HPS} & \textbf{Edge HPS} & $\boldsymbol{\Delta}$ \textbf{Step} \\
% \midrule
% 2-hop & 1252 & 0.918 & 0.780 & 0.049 & 0.921 & 0.820 & 0.133 & 0.973 & 0.920 & 0.159 \\
% 3-hop & 760 & 0.716 & 0.480 & -0.622 & 0.842 & 0.638 & -0.041 & 0.962 & 0.882 & 0.217 \\
% 4-hop & 405 & 0.590 & 0.340 & -1.062 & 0.721 & 0.478 & -0.397 & 0.917 & 0.814 & 0.165 \\\midrule
% simple\_tree & 1820 & 0.855 & 0.698 & -0.162 & 0.897 & 0.780 & 0.069 & 0.968 & 0.910 & 0.169 \\
% v\_structure & 351 & 0.683 & 0.350 & -0.723 & 0.792 & 0.454 & -0.108 & 0.957 & 0.836 & 0.393 \\
% complex\_tree & 246 & 0.556 & 0.349 & -1.193 & 0.714 & 0.511 & -0.459 & 0.907 & 0.821 & -0.061 \\
% \bottomrule
% \end{tabular}}
% \caption{Setting~C end-to-end graph generation performance stratified by hop count and topology slices.}
% \label{tab:topology-strat}
% \end{table*}

Setting~C reveals that end-to-end graph generation difficulty depends on both hop count and topology. We categorize MuSiQue gold dependency DAGs into three topology buckets: \textit{simple\_tree} (no merge node), \textit{v\_structure} (one merge node with parallel parents), and \textit{complex\_tree} (deeper, larger, or multi-merge graphs); formal criteria appear in Appendix~\ref{app:within-hop-topology}.

As shown in Table~\ref{tab:topology-strat}, HPS and Edge HPS decline monotonically from 2-hop to 4-hop across all three anchor models; the drop is steepest for Llama-3.1-70B, whose Edge HPS falls from 0.780 at 2-hop to 0.340 at 4-hop. Topology provides an additional diagnostic axis: \textit{simple\_tree} is consistently easier than merge or deeper structures across all models, but the relative difficulty of \textit{v\_structure} versus \textit{complex\_tree} is not fixed. For Qwen3-32B, \textit{complex\_tree} Edge HPS (0.511) exceeds \textit{v\_structure} (0.454), while for Llama-3.1-70B the two are effectively equal (0.350 vs.\ 0.349).
Within-hop comparisons reveal that topology shapes the \emph{type} of failure rather than only its degree: merge structures lower Edge HPS without reducing step coverage, whereas deeper structures induce under-decomposition (Appendix~\ref{app:within-hop-topology}).
The per-type Setting~C breakdown and full A/B/C model results for 2WikiMultiHopQA appear in Appendix~\ref{app:twowiki-stage1}.

% ============================================================
\subsection{Failure Modes Across Diagnostic Settings}
\label{sec:stage1-error-modes}
% ============================================================

Table~\ref{tab:stage1-error-summary} reports the dominant non-aligned error type for each model and setting; Table~\ref{tab:error-tag-brief} defines the tags that appear in it. Full taxonomy and full-dev error counts appear in Appendix~\ref{app:stage1-error-diagnostics}. Together, they show a directional shift in the failure bottleneck: as gold scaffolding is removed, errors do not merely increase within the same graph category, but move earlier in the construction pipeline, from edge wiring, to local node grounding, to decomposition itself.

% \vspace{0.15cm}
\begin{table}[htbp]
\centering
\small
\setlength{\tabcolsep}{4pt}
\begin{tabular}{lp{0.68\linewidth}}
\toprule
\textbf{Tag} & \textbf{Description} \\
\midrule
\multicolumn{2}{l}{\textit{Setting A:edge wiring failures}} \\
miss.\ terminal  & Gold dependency into $q_{\mathrm{final}}$ is omitted. \\
shortcut repl.\  & Gold dependency replaced by a shortcut edge, often into $q_{\mathrm{final}}$. \\
broad mixed      & Multiple answers, evidence, and edge failures with no dominant cause. \\
\midrule
\multicolumn{2}{l}{\textit{Setting B:local grounding failures}} \\
ans.+evid.\      & Both generated answer and evidence attribution fail at a fixed node. \\
\midrule
\multicolumn{2}{l}{\textit{Setting C:decomposition failures}} \\
under-decomp.\   & Predicted graph has fewer steps than gold. \\
struct.\ gap     & Steps align to gold but their dependency structure is wrong. \\
evid.\ gap       & Steps and structure align but evidence attribution fails. \\
\bottomrule
\end{tabular}
\caption{Error tags appearing in Table~\ref{tab:stage1-error-summary}.}
\label{tab:error-tag-brief}
\end{table}

\begin{table*}[t]
\centering
\small
\begin{tabular}{lccc}
\toprule
\textbf{Model} & \textbf{Setting~A} & \textbf{Setting~B} & \textbf{Setting~C} \\
\midrule
Llama-3.1-8B         & miss.\ terminal 45.7\% & broad mixed 35.0\% & struct.\ gap 37.9\%  \\
Qwen3-8B             & miss.\ terminal 46.5\% & ans.+evid.\ 40.7\% & under-decomp.\ 38.6\% \\
Qwen3-14B            & shortcut repl.\ 7.1\%  & ans.+evid.\ 34.0\% & under-decomp.\ 33.7\% \\
Qwen3-32B            & shortcut repl.\ 3.2\%  & ans.+evid.\ 30.2\% & struct.\ gap 19.5\%  \\
Qwen3-32B-Reasoning  & shortcut repl.\ 1.8\%  & ans.+evid.\ 25.9\% & under-decomp.\ 20.7\% \\
Llama-3.1-70B        & shortcut repl.\ 19.2\% & ans.+evid.\ 23.7\% & under-decomp.\ 34.7\% \\
DeepSeek V4 Flash    & shortcut repl.\ 1.9\%  & ans.+evid.\ 24.5\% & under-decomp.\ 20.9\% \\
GPT-5                & shortcut repl.\ 2.8\%  & ans.+evid.\ 24.2\% & struct.\ gap 17.5\%  \\
Claude Sonnet~4.5    & shortcut repl.\ 1.3\%  & ans.+evid.\ 18.6\% & evid.\ gap 18.7\%    \\
\bottomrule
\end{tabular}
\caption{Top non-aligned error type and rate per model per diagnostic setting (MuSiQue full dev set). Each entry is the most frequent error category excluding \texttt{fully\_aligned}. }
\label{tab:stage1-error-summary}
\end{table*}

The dominant error changes with the part of graph construction that is left to the model. In Setting~A, node content is fixed, so failures are restricted to edge wiring. The main split is between disconnected and shortcuted graphs: the 8B models most often omit the terminal dependency into $q_{\mathrm{final}}$, producing \texttt{missing\_terminal\_edge} above 45\% (partially schema-artifact induced; see Appendix~\ref{app:schema-sensitivity}), whereas Llama-3.1-70B more often preserves connectivity by replacing an intermediate dependency with a direct edge to $q_{\mathrm{final}}$ (\texttt{shortcut\_replacement}, 19.2\%; Qwen3-32B: 2.8\%). This suggests that stronger models may produce structurally plausible but logically compressed graphs, while smaller models are more prone to format-level or connectivity failures.
In Setting~B, the decomposition remains fixed, but the model must generate node answers and evidence. The bottleneck therefore shifts from topology to local grounding: \texttt{answer\_and\_evidence} co-failure becomes the dominant category for all models except Llama-3.1-8B. In Setting~C, the decomposition scaffold is removed, and failures move further upstream. \texttt{under\_decompose} or \texttt{structure\_gap} is the top error for eight of nine models, exceeding 33\% for the four weakest models.
\section{Stage~2: Uncertainty Quantification with Reasoning Graphs}
\label{sec:sc}

Stage~2 evaluates how reasoning graphs affect confidence estimation for multi-hop QA final answers. We first use the gold MuSiQue graph to isolate graph use from graph construction errors: if the structure is correct, does it help more as prompt context or as an explicit computation structure? We then test how upstream construction errors and graph complexity affect the results.

\subsection{Methods}
\label{sec:stage2-setup}
\label{sec:stage2-methods}

\paragraph{Setup.}
The main experiments use Llama-3.1-70B on MuSiQue; cross-model results for Qwen3-32B and a cross-dataset diagnostic on 2WikiMultiHopQA appear in Appendix~\ref{app:stage2-cross-model} and~\ref{app:stage2-twowiki}. Sampling-based methods draw $K{=}10$ samples at temperature~0.7; single-call verbalized methods use one completion per question. Answer quality is measured by EM and F1 under SQuAD-style normalization \citep{rajpurkar2016squad}. Confidence quality is measured by AUROC \citep{hanley1982meaning}, Expected Calibration Error~(ECE) \citep{naeini2015obtaining, guo2017calibration}, and Brier score \citep{glenn1950verification}, with exact-match correctness as the binary target. For answer clustering, we apply SQuAD-style string normalization to model predictions only; gold answers and aliases are not used for clustering and enter only official EM/F1 scoring. Prompt details appear in Appendix~\ref{app:prompts}; call-budget statistics appear in Appendix~\ref{app:call-budget}.

\paragraph{Baselines.}
We compare seven methods grouped into four categories: \textit{graph-free sampling}, \textit{graph-as-prompt}, \textit{verbalized confidence}, and \textit{graph-as-computation}. Graph-free sampling includes Direct-SC, which samples final answers without a reasoning trace, and CoT-SC, which samples chain-of-thought traces \citep{wang2022self}; for both, confidence is the normalized frequency of the majority answer cluster. We introduce Decomp-SC, which provides gold sub-questions as context, and Graph-CoT-SC, which additionally provides gold dependency edges representing graph-as-prompt. Verbalized confidence includes CoT-Conf, a single-pass chain-of-thought baseline that elicits a scalar confidence score in $[0,1]$ directly, and vPGM-Style, our adaptation of verbalized probabilistic graphical modeling to open-ended multi-hop QA \citep{huang_bayesagent_2026,tanneru_quantifying_2023}; implementation details appear in Appendix~\ref{app:vpgm}.

\paragraph{Sampled BN.}
We also introduce Sampled~BN representing the graph-as-computation category. Rather than using the graph as prompt context or eliciting verbalized probabilities, it uses the dependency DAG as an explicit computation structure.
Each node $v_i$ is treated as a discrete answer-state variable, whose states are empirical normalized-answer clusters.
For each observed parent-state assignment, we sample the child node $K$ times and estimate the local conditional $\hat{p}(a_i \mid \mathbf{a}_{\mathrm{pa}(i)})$ by empirical frequency over the $K$ samples, where lowercase letters denote realized values of $A_i$ and $A_{\mathrm{pa}(i)}$; root-node distributions are estimated analogously without parent conditions.
Let $f$ denote the final-answer node, $y$ a candidate answer string, and $\mathbf{a}_{V \setminus \{f\}}$ a joint assignment of answer states to all non-final nodes, ranging over sampled answer states. Final-answer confidence is:
\[
\hat{p}(A_f=y)
=
\sum_{\mathbf{a}_{V \setminus \{f\}}}
\prod_{v_i \in V}
\hat{p}\!\left(a_i \mid \mathbf{a}_{\mathrm{pa}(i)}\right),
\]
where $a_f = y$.
This computation is an empirical BN-style propagation over model-sampled answer clusters, not an estimate of the true posterior distribution.
% This differs from CoT-SC, which estimates confidence from repeated complete-answer samples without explicit parent-conditioned node states.

% ============================================================
\subsection{Gold-Graph Results: Prompt Context or Computation?}
\label{sec:stage2-main}
\label{sec:stage2-verbalization}
% ============================================================

% Table: Stage 2 confidence estimation — main results.
% n=2417 for all methods.
% Columns: Answer EM/F1, AUROC EM, ECE EM, Brier EM, Avg Conf.
% Bold marks per-column best.
\begin{table}[htbp]
\centering
\small
\setlength{\tabcolsep}{5pt}
\resizebox{\columnwidth}{!}{%
\begin{tabular}{l r r r r r}
\toprule
\textbf{Method}
  & \textbf{EM}$\uparrow$
  & \textbf{F1}$\uparrow$
  & \textbf{AUROC}$\uparrow$
  & \textbf{ECE}$\downarrow$
  & \textbf{Brier}$\downarrow$ \\
\midrule
Direct-SC
  & 0.462 & 0.569 & 0.718 & 0.325 & 0.319 \\
CoT-Conf
  & 0.487 & 0.593 & 0.686 & 0.271 & 0.297 \\
CoT-SC
  & \textbf{0.572} & 0.680 & 0.766 & \textbf{0.161} & \textbf{0.220} \\
Decomp-SC
  & 0.550 & 0.659 & 0.770 & 0.234 & 0.247 \\
Graph-CoT-SC
  & 0.562 & 0.671 & 0.764 & 0.238 & 0.251 \\
vPGM-Style
  & 0.439 & 0.550 & 0.622 & 0.235 & 0.295 \\
Sampled BN
  & 0.568 & \textbf{0.691} & \textbf{0.795} & 0.206 & 0.225 \\
\bottomrule
\end{tabular}}
\caption{Stage~2 confidence estimation on MuSiQue full dev set.}
\label{tab:stage2-main}
\end{table}

% \vspace{-0.15cm}

\begin{table*}[!t]
\centering
\small
\setlength{\tabcolsep}{4pt}
\resizebox{\textwidth}{!}{
\begin{tabular}{l r r G D r G D r G D r G D r G D}
\toprule
\textbf{Method} & \textbf{N}
& \multicolumn{3}{c}{\textbf{EM}}
& \multicolumn{3}{c}{\textbf{F1}}
& \multicolumn{3}{c}{\textbf{AUROC}}
& \multicolumn{3}{c}{\textbf{ECE $\downarrow$}}
& \multicolumn{3}{c}{\textbf{Brier $\downarrow$}} \\
\cmidrule(lr){3-5}
\cmidrule(lr){6-8}
\cmidrule(lr){9-11}
\cmidrule(lr){12-14}
\cmidrule(lr){15-17}
& & \textbf{Gen} & \textbf{Gold} & $\boldsymbol{\Delta}$
& \textbf{Gen} & \textbf{Gold} & $\boldsymbol{\Delta}$
& \textbf{Gen} & \textbf{Gold} & $\boldsymbol{\Delta}$
& \textbf{Gen} & \textbf{Gold} & $\boldsymbol{\Delta}$
& \textbf{Gen} & \textbf{Gold} & $\boldsymbol{\Delta}$ \\
\midrule
Graph-CoT-SC & 446
& 0.476 & 0.469 & +0.007
& 0.620 & 0.622 & -0.002
& 0.689 & 0.714 & -0.025
& 0.313 & 0.324 & -0.011
& 0.326 & 0.322 & +0.004 \\
Sampled BN & 446
& 0.459 & 0.493 & -0.034
& 0.599 & 0.657 & -0.058
& 0.738 & 0.755 & -0.017
& 0.300 & 0.283 & +0.017
& 0.300 & 0.285 & +0.015 \\
\bottomrule
\end{tabular}
}
\caption{
Generated-edge stress test on valid non-exact Setting~A graphs (446 examples).
Gen: performance under Llama-3.1-70B generated edges.
Gold: performance using gold edges on the same examples.
$\Delta$ = Gen $-$ Gold.
}
\label{tab:generated-edge-stress}
\end{table*}

Table~\ref{tab:stage2-main} reports the results under the gold-graph condition. \textit{Verbalized confidence} performs weakest: vPGM-Style and CoT-Conf reach AUROC~0.622 and 0.686, both below the graph-free Direct-SC baseline at 0.718. \textit{Graph-as-prompt} provides little ranking benefit over CoT-SC: Decomp-SC and Graph-CoT-SC reach AUROC 0.770 and 0.764 versus 0.766 for CoT-SC, while both have lower EM.

The central finding is a metric-specific split between \textit{graph-free sampling} and \textit{graph-as-computation}. CoT-SC leads on answer accuracy and calibration with EM~0.572, ECE~0.161, Brier~0.220 while Sampled~BN significantly improves confidence ranking, with AUROC 0.795 versus 0.766 for CoT-SC, while achieving comparable F1 (0.691 vs. 0.680). Neither method dominates across all metrics: graph-as-computation improves ranking at the cost of calibration. A smoothing sensitivity analysis shows that this calibration degradation is
partly attributable to sparse empirical CPT estimates, while answer quality
and confidence ranking remain largely unchanged
(Appendix~\ref{app:smoothing}).
Cross-model results suggest that gains from graph-based propagation depend on the
quality of the underlying node-level predictions, with the improvement over CoT-SC
observed for Llama-3.1-70B but not for Qwen3-32B
(Appendix~\ref{app:stage2-cross-model}).
Because AUROC summarizes confidence ranking, we also report selective-prediction curves in Appendix~\ref{app:selective} and pairwise correctness contingency between Sampled~BN and CoT-SC in Appendix~\ref{app:stage2-contingency}. Selective-prediction curves show that Sampled~BN's ranking advantage concentrates at intermediate coverage, while its near-symmetric disagreement with CoT-SC indicates complementary rather than uniformly better predictions (Appendix~\ref{app:selective} and~\ref{app:stage2-contingency}). We also assess the statistical significance of all comparative claims in Sections \ref{sec:stage2-main}–\ref{sec:stage2-boundary} using paired bootstrap tests; full results are reported in Appendix~\ref{app:significance}.

\begin{figure}[htbp]
\centering
\includegraphics[width=\linewidth]{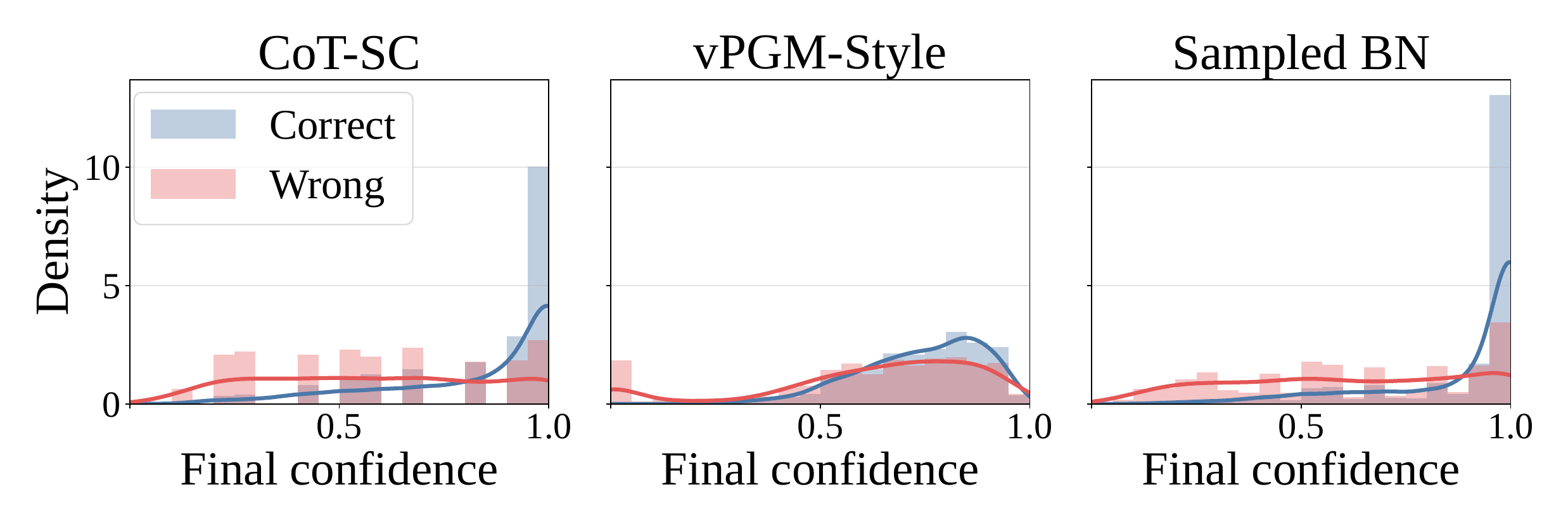}
\caption{Confidence distributions for CoT-SC, vPGM-Style, and Sampled~BN, separated by final-answer correctness.}
\label{fig:stage2-confidence-dist}
\end{figure}

\paragraph{Confidence distributions.}
Figure~\ref{fig:stage2-confidence-dist} shows that the methods induce different confidence shapes.
The distributions explain why ranking and calibration diverge. CoT-SC has moderate separation but many predictions concentrate around similar confidence values. vPGM-Style is less discriminative because correct and wrong predictions overlap heavily in the mid-to-high range. Sampled BN produces the sharpest separation for many correct predictions, but its wrong predictions also include a high-confidence tail. Thus, a sharper graph-propagated signal improves ranking while creating calibration risk when the propagated evidence is wrong.

% \vspace{-0.15cm}

\paragraph{Node-level diagnostics.}
\label{sec:stage2-per-node}
Sampled~BN exposes node-level confidence through its sampling and graph-propagation procedure, enabling diagnostics that final-answer confidence methods cannot provide. We first evaluate whether node confidence identifies incorrect intermediate answers. Node-level AUROC is 0.744 over all evaluated nodes and 0.745 on examples with wrong final answers, showing that node confidence is informative beyond the final prediction alone.
We then test root-cause localization on wrong-answer examples. We define the root-cause node as the first node in topological order whose generated answer is wrong while all of its parent nodes are correct. Intermediate nodes are ranked by ascending confidence, with the most uncertain node ranked first. This simple ranking achieves Top-1 accuracy 0.700 and Mean Reciprocal Rank (MRR) 0.836, indicating that the earliest erroneous node is usually placed near the top of the uncertainty ranking. Positional breakdowns appear in Appendix~\ref{app:stage2-per-node-full}.

% ============================================================
\subsection{Generated Edge Results: How Does it Affect Performance?}
\label{sec:generated-edge-stress}
% ============================================================

The gold-graph comparison isolates how structure can be used when the graph is correct; we next test what happens when the edge structure comes from Stage~1 generation. We use Llama-3.1-70B Setting~A outputs as generated structures: gold nodes are fixed, and only dependency edges are replaced. We retain only structurally valid generated graphs, requiring legal endpoints, acyclicity, a valid topological order, no outgoing edge from $q_{\text{final}}$, and reachability from every non-final node to $q_{\text{final}}$. This yields a valid generated graph rate of 96.3\%. The compact main table focuses on valid non-exact graphs.

We evaluate Graph-CoT-SC and Sampled~BN because both explicitly depend on the edge set. Graph-CoT-SC uses edges as prompt context, whereas Sampled~BN uses edges operationally to define parent sets and conditional factorization. Table~\ref{tab:generated-edge-stress} compares generated and gold edges on the 446 valid non-exact examples. Graph-CoT-SC is largely robust: F1 changes by only $-0.002$ (0.620 vs.\ 0.622). Sampled~BN is more sensitive: F1 drops from 0.657 to 0.599 ($-5.8$ percentage points, pp), AUROC from 0.755 to 0.738 ($-1.7$~pp), and both ECE and Brier worsen. This qualifies the gold-graph results: graph-as-computation can provide useful confidence signals under reliable structure, but it also inherits upstream edge errors more directly than graph-as-prompt. Importantly, this degradation remains after filtering out structurally invalid graphs, indicating that legal DAG structure alone is insufficient for reliable graph-based computation.

% ============================================================
\subsection{Sensitivity to Graph Structure}
\label{sec:stage2-boundary}
% ============================================================

Whether Sampled~BN improves over CoT-SC depends on graph structure.
Table~\ref{tab:stage2-structure-compact} shows that Sampled~BN leads in F1 on 2-hop ($+2.5$~pp) and 3-hop ($+3.3$~pp) but falls behind on 4-hop ($-7.8$~pp).
The same pattern holds by topology: Sampled~BN leads on \textit{simple\_tree} ($+2.4$~pp) and \textit{v\_structure} ($+3.6$~pp), but drops below CoT-SC on \textit{complex\_tree} ($-12.3$~pp).
AUROC shows smaller reversals: Sampled~BN loses only $2.8$~pp on 4-hop and is effectively tied on \textit{complex\_tree} ($-0.4$~pp), retaining a positive margin on all other slices.

This pattern is consistent with a propagation failure mode: in deeper or more nested graphs, incorrect intermediate answer states can become parent states for downstream conditionals, causing local errors to be carried through the computation. 
Thus, graph-as-computation provides useful confidence signals under favorable structure, but its answer quality is more exposed to intermediate-state errors in complex graphs.
Full results for all seven methods stratified by hop count and topology appear in Appendix~\ref{app:stage2-stratified-full}.

% Compact Stage 2 structure-stratified comparison for the main text.
\vspace{0.15cm}
\begin{table}[htbp]
\centering
\resizebox{\columnwidth}{!}{%
\begin{tabular}{l rr rr}
\toprule
\textbf{Slice}
  & \multicolumn{2}{c}{\textbf{F1}}
  & \multicolumn{2}{c}{\textbf{AUROC}} \\
\cmidrule(lr){2-3}\cmidrule(lr){4-5}
  & \textbf{CoT-SC} & \textbf{Sampled BN}
  & \textbf{CoT-SC} & \textbf{Sampled BN} \\
\midrule
2-hop & 0.717 & \textbf{0.742} & 0.753 & \textbf{0.795} \\
3-hop & 0.673 & \textbf{0.706} & 0.770 & \textbf{0.816} \\
4-hop & \textbf{0.577} & 0.499 & \textbf{0.780} & 0.752 \\
\midrule
simple\_tree & 0.713 & \textbf{0.737} & 0.761 & \textbf{0.789} \\
v\_structure & 0.519 & \textbf{0.555} & 0.761 & \textbf{0.790} \\
complex\_tree & \textbf{0.661} & 0.538 & \textbf{0.796} & 0.792 \\
\bottomrule
\end{tabular}}
\caption{Structure-stratified F1 and AUROC for CoT-SC and Sampled~BN. Bold marks the higher value per row.}
\label{tab:stage2-structure-compact}
\end{table}

\vspace{-0.3cm}
\section{Conclusion}
\label{sec:conclusion}

We presented a two-stage diagnostic framework for evaluating LLM-generated reasoning graphs. The results suggest a conditional answer: strong models can produce reliable dependency structures in controlled settings, but graph reliability is not automatic. It varies with model family, schema design, structural complexity, and the amount of gold scaffolding available. The failure bottleneck shifts systematically as gold scaffolding is removed. These findings argue for treating reasoning graphs as intermediate computational objects that require direct evaluation. A natural next step is developing methods that degrade gracefully when graph quality varies, combining graph-structured and graph-free confidence signals to preserve gains while limiting the damage from imperfect graphs.
% \newpage
\section{Limitations}
\label{sec:limitations}

This study is diagnostic rather than a complete end-to-end graph-reasoning system. Stage~1 evaluates graph construction under controlled settings on MuSiQue and a stratified 2WikiMultiHopQA sample, so the specific topology and error-mode profiles may change on datasets with different decomposition conventions, evidence distributions, or question templates. Moreover, gold decompositions are treated as reference structures, although some multi-hop questions may admit multiple valid reasoning graphs. Our semantic node matching and reachability-based Edge HPS tolerate some
differences in wording and decomposition granularity, but they cannot capture
all alternative valid reasoning structures.

The Stage~2 methods also differ in call budget and output interface. Sampled~BN uses graph-conditioned node sampling and is not compute-matched to single-call verbalized confidence or self-consistency baselines. It also relies on finite empirical answer-state supports and a BN-style conditional factorization, which may be misspecified when answer states have dependencies not captured by the annotated graph. These choices make the comparisons best interpreted as controlled design diagnostics rather than deployment-ready method rankings.

\section*{Acknowledgments}

ChatGPT was used to assist with language editing and manuscript
revision, and Claude Code was used for code editing and debugging.
All generated suggestions were reviewed and verified by the authors,
who take full responsibility for the paper and the released artifacts.

\bibliography{reasoning}

% \clearpage
\appendix
% \onecolumn

\section{Prompt Templates}
\label{app:prompts}

This appendix documents the prompt templates for Stage~1 graph construction and Stage~2 uncertainty quantification. We use publicly released research datasets and model APIs/checkpoints under their respective licenses or access terms, and do not redistribute the original datasets or model weights.

\subsection{Stage~1: Graph-Construction Prompts}
\label{app:stage1-prompts}

Stage~1 uses three prompt versions corresponding to Settings~A, B, and C. All use temperature~0; Qwen3 models are called with \texttt{enable\_thinking=false}. Paragraphs are formatted as \texttt{[idx N] Title: \{title\}\textbackslash n\{text\}}, one block per entry separated by blank lines.

\paragraph{System prompt (shared).}

\begin{quote}
\ttfamily
You are a careful graph induction assistant.
Follow the instructions exactly and output valid JSON only.
\end{quote}

\paragraph{Shared constraint: Immediate Edge Clarification.}

All three settings include the following block verbatim to enforce transitive reduction:

\begin{lstlisting}
IMMEDIATE EDGE CLARIFICATION:
Only predict immediate prerequisite edges.
An edge qi -> qj means that the answer to qi is directly needed to
answer qj.
Do not output transitive edges.
If information flows qi -> qj -> qk, do not also add qi -> qk unless
qk directly requires qi independently of qj.
Do not output broad final-answer support edges.
Do not connect qi directly to q_final if qi is only used through
another intermediate node qj.
For example, if q2 requires q1 and q_final is answered by q2, output
q1->q2 and q2->q_final, not q1->q_final.
The target graph is the transitive-reduced dependency DAG, not the
transitive closure of all support relations.
\end{lstlisting}

\paragraph{Shared constraint: Confidence Semantics.}

\begin{lstlisting}
CONFIDENCE SEMANTICS:
For each edge, provide a confidence value in [0, 1] representing your
estimate of the probability that this edge exists in the correct
dependency graph.
- 0.9-1.0: almost certain the dependency exists
- 0.7-0.9: likely but some uncertainty
- 0.5-0.7: ambiguous - include only if the edge is more likely than not
- below 0.5: do not include this edge at all
\end{lstlisting}

Settings~B and C additionally specify per-node confidence fields (\texttt{answer\_confidence}, \texttt{attribution\_confidence}) with the same bin semantics.

\paragraph{Setting~A: Edge-only prediction.}

The model receives the main question, candidate paragraphs (gold plus distractors), and gold reasoning nodes with their sub-questions, sub-answers, and supporting paragraph IDs already filled in. The task is limited to predicting dependency edges and topological order.

\begin{lstlisting}
You are given a multihop question, candidate paragraphs (each labeled
with an integer idx), and fixed gold reasoning nodes. Each fixed node
already includes the gold sub-question, gold sub-answer, and gold
supporting paragraph id(s).

Your only task is to predict the directed dependency edges among the
fixed nodes and provide a topological order.

CONSTRAINTS:
- Do NOT output nodes.
- Do NOT output sub_answers.
- Do NOT output supporting_paragraph_ids or evidence ids.
- Do NOT output node_confidences.
- Use only the fixed node ids shown below as edge endpoints.
- A directed edge qi -> qj means "answering qj requires knowing the
  answer to qi".
- The graph must be acyclic.
- If you include topological_order, it should contain every fixed node
  id exactly once. If omitted or invalid, it will be recovered from
  the edge DAG.
- q_final should be the sink node and should not have outgoing edges.
- Every non-final node should have a directed path to q_final.
- Edge justification is optional; include it when useful.
- Output only edges and topological_order.
\end{lstlisting}

\paragraph{Setting~B: Fixed decomposition.}

The model receives the main question, candidate paragraphs (including distractors), and gold sub-question texts. It must generate sub-answers, edge structure, evidence attribution, and confidence scores.

\begin{lstlisting}
You are given a multihop question, candidate paragraphs, some of which
may be irrelevant distractors (each labeled with an integer idx), and
a fixed list of reasoning node questions. Each node's sub_question
text is fixed, but you must generate the corresponding sub_answer
yourself.

Your task is to:
(1) fill in a sub_answer for every given node, including q_final,
(2) identify the dependency structure among these fixed nodes,
(3) attribute each node to the paragraph(s) that support its answer,
(4) provide calibrated confidence for each edge, each node's answer,
    and each node's paragraph attribution.

CONSTRAINTS:
- Do NOT invent new nodes.
- Do NOT modify the given node ids.
- Do NOT modify the given sub_question texts.
- You must generate sub_answer for every node, including q_final.
- The output must include exactly the given nodes.
- A directed edge qi -> qj means "answering qj requires knowing the
  answer to qi".
- The graph must be acyclic.
- q_final should be the sink node and should not have outgoing edges.
- Every non-final node should have a directed path to q_final.
- For supporting_paragraph_ids, use only the integer idx values from
  the given paragraph list. For q_final, always use
  supporting_paragraph_ids: [].
\end{lstlisting}

\paragraph{Setting~C: Free decomposition.}

The model receives only the main question and candidate paragraphs (including distractors). It must generate the full graph from scratch.

\begin{lstlisting}
You are given a multihop question and candidate paragraphs, some of
which may be irrelevant distractors (each labeled with an integer
idx). Your task is to:
(1) decompose the main question into a set of atomic reasoning nodes;
    the dependency edges and topological_order will define their
    reasoning order, where each node is a (sub-question, sub-answer)
    pair,
(2) identify the dependency structure among these nodes,
(3) attribute each node to the paragraph(s) that support its answer,
(4) provide calibrated confidence for each edge, each node's answer,
    and each node's paragraph attribution.

CONSTRAINTS:
- Each sub-question must be atomic: answerable by a single fact lookup
  or single-step inference.
- Each node must contain both sub_question and sub_answer.
- Include a final node q_final where sub_question is the original main
  question and sub_answer is the final answer.
- A directed edge qi -> qj means "answering qj requires knowing the
  answer to qi".
- The graph must be acyclic.
- Aim for decomposition granularity similar to standard multihop QA
  decomposition (typically 2-5 sub-questions for 2-4-hop questions).
- q_final should be the sink node and should not have outgoing edges.
- Every non-final node should have a directed path to q_final.
- For supporting_paragraph_ids, use only the integer idx values from
  the given paragraph list. For q_final, always use
  supporting_paragraph_ids: [].
\end{lstlisting}

\paragraph{Output schemas.}

Setting~A uses an edge-only schema:

\begin{lstlisting}
{
  "edges": [
    {"from": "q1", "to": "q2",
     "confidence": 0.85,
     "justification": "q2 directly requires q1 because ..."}
  ],
  "topological_order": ["q1", "q2", "q_final"]
}
\end{lstlisting}

Settings~B and C share the full graph schema:

\begin{lstlisting}
{
  "nodes": [
    {"id": "q1", "sub_question": "...", "sub_answer": "...",
     "supporting_paragraph_ids": [3]},
    {"id": "q_final", "sub_question": "...", "sub_answer": "...",
     "supporting_paragraph_ids": []}
  ],
  "edges": [
    {"from": "q1", "to": "q2",
     "confidence": 0.85,
     "justification": "q2 directly requires q1 because ..."}
  ],
  "node_confidences": [
    {"id": "q1", "answer_confidence": 0.95,
     "attribution_confidence": 0.9}
  ],
  "topological_order": ["q1", "q2", "q_final"]
}
\end{lstlisting}

All schemas are preceded by: \textit{``The JSON schema below is a field-format example only; do not copy its example edge pattern as a topology template.''}

\paragraph{Schema example variants (E1--E5).}

All three settings embed a concrete schema example in the prompt. Appendix~\ref{app:schema-sensitivity} shows that the choice of example affects parse behavior: E1 uses a concrete multi-node example topology that small models mimic, producing disconnected graphs at a 47.2\% rate; E3 replaces the concrete nodes with neutral placeholders, reducing the disconnected-graph rate to below 2\%. Full variant definitions and five-model comparisons appear in Appendix~\ref{app:schema-full-dev}.

\subsection{Stage~2: Uncertainty Quantification Prompts}
\label{app:stage2-prompts}

All Stage~2 methods receive the main question and the full MuSiQue paragraph context. Sampling-based methods draw $K{=}10$ samples at temperature~0.7; verbalized methods (CoT-Conf, vPGM-Style) use one call per example.

\paragraph{Shared system prompt (Direct-SC, CoT-SC, Decomp-SC, Graph-CoT-SC).}

\begin{lstlisting}
You are answering a multi-hop question using only the provided
paragraphs as factual sources.

Follow the requested output format exactly.
The final answer must be a short phrase, named entity, date, number,
or title.
Do not include citations in the final answer.
\end{lstlisting}

\paragraph{Direct-SC user prompt.}

\begin{lstlisting}
=== Question ===
{question}

=== Candidate Paragraphs ===
{paragraphs_block}

Answer the question using only the candidate paragraphs.

Output exactly one line:
FINAL: <short answer>
\end{lstlisting}

\paragraph{CoT-SC user prompt.}

\begin{lstlisting}
=== Question ===
{question}

=== Candidate Paragraphs ===
{paragraphs_block}

Think through the question step by step using only the candidate
paragraphs. Then give the final answer as a short phrase.

Output in exactly this format:

REASONING:
<brief step-by-step reasoning>

FINAL: <short answer>
\end{lstlisting}

\paragraph{Decomp-SC user prompt.}

\begin{lstlisting}
=== Question ===
{question}

=== Candidate Paragraphs ===
{paragraphs_block}

=== Given Sub-questions ===
{node_list_without_edges}

You are given sub-questions that may help answer the original
question. Use only the candidate paragraphs. Answer each sub-question
briefly, then answer the original question.

Do not infer or output dependency edges.
Do not add new sub-questions.

Output exactly one line for each given sub-question, followed by
q_final.

Required format:
q1: <short answer>
q2: <short answer>
...
q_final: <short answer to the original question>

No extra text.
\end{lstlisting}

\paragraph{Graph-CoT-SC user prompt.}

This template is used for both the gold-graph and generated-edge conditions (\S\ref{sec:generated-edge-stress}); the difference is whether \texttt{\{edge\_list\}} is rendered from gold edges or from Stage~1-generated edges.

\begin{lstlisting}
=== Question ===
{question}

=== Candidate Paragraphs ===
{paragraphs_block}

=== Given Reasoning Nodes ===
{node_list}

=== Given Dependency Edges ===
{edge_list}

The nodes and edges define a reasoning scaffold.
An edge qi -> qj means that answering qj directly depends on the
answer to qi.

Use the given scaffold as a reasoning guide.
Use only the candidate paragraphs.
Answer each node briefly, then answer q_final.

Do not add new nodes.
Do not add new edges.
Do not change the given edges.
Do not output explanations beyond the required answer lines.

Output exactly one line for each given node.

Required format:
q1: <short answer>
q2: <short answer>
...
q_final: <short answer to the original question>

No extra text.
\end{lstlisting}

\paragraph{CoT-Conf system prompt.}

\begin{lstlisting}
You are answering a multi-hop question using only the provided
candidate paragraphs as factual sources.
Think step by step, then provide a final short answer and a confidence
score. The confidence should be a number between 0 and 1 indicating
how likely your final answer is correct.
Output valid JSON only.
\end{lstlisting}

\paragraph{CoT-Conf user prompt.}

\begin{lstlisting}
=== Question ===
{question}

=== Candidate Paragraphs ===
{paragraphs_block}

Think step by step and answer the question.

Return valid JSON only in this format:
{
  "reasoning": "<concise step-by-step reasoning based only on the
                paragraphs>",
  "final_answer": "<short answer>",
  "confidence": <number between 0 and 1>
}
\end{lstlisting}

\paragraph{vPGM-Style system prompt.}

\begin{lstlisting}
You are answering a multi-hop question using only the provided
candidate paragraphs as factual sources.

You will perform verbalized probabilistic graphical modeling.
Estimate each probability carefully as a number between 0 and 1.
Use the latent variables to assess uncertainty before giving the
final answer.

Output valid JSON only.
\end{lstlisting}

\paragraph{vPGM-Style user prompt.}

Three latent variables are elicited before the final answer: $Z_E$ (evidence coverage), $Z_D$ (distractor control), and $Z_R$ (reasoning validity). The model estimates each conditional probability in sequence, then produces the final answer.

\begin{lstlisting}
=== Question ===
{question}

=== Candidate Paragraphs ===
{paragraphs_block}

We use the following latent variables for multi-hop question answering:

Z_E: Evidence Coverage. Whether the candidate paragraphs contain
sufficient evidence for all necessary reasoning steps.

Z_D: Distractor Control. Whether irrelevant or misleading distractor
paragraphs can be ignored.

Z_R: Reasoning Validity. Whether the evidence can be connected into
a valid multi-hop reasoning chain without logical gaps.

Perform step-by-step probabilistic reasoning:

1. Estimate P(Z_E | X): assess whether the paragraphs contain enough
   evidence for all required hops.

2. Estimate P(Z_D | Z_E, X): assess whether the reasoning process can
   avoid irrelevant or misleading distractor paragraphs.

3. Estimate P(Z_R | Z_E, Z_D, X): assess whether a valid multi-hop
   reasoning chain can be formed.

4. Estimate P(Y = y_hat | Z_E, Z_D, Z_R, X): generate a final short
   answer y_hat, then estimate the probability that this specific
   answer is correct.

Return valid JSON only in this format:

{
  "P_Z_E_given_X": {
    "probability": <number between 0 and 1>,
    "rationale": "<brief rationale>"
  },
  "P_Z_D_given_Z_E_X": {
    "probability": <number between 0 and 1>,
    "rationale": "<brief rationale>"
  },
  "P_Z_R_given_Z_E_Z_D_X": {
    "probability": <number between 0 and 1>,
    "rationale": "<brief rationale>"
  },
  "final_answer": "<short answer>",
  "P_Y_given_Z_E_Z_D_Z_R_X": {
    "probability": <number between 0 and 1>,
    "rationale": "<brief rationale>"
  }
}
\end{lstlisting}

\paragraph{Sampled~BN: root-node prompt.}

Sampled~BN uses separate prompts for root and non-root nodes; both are sampled $K{=}10$ times at temperature~0.7.

\begin{lstlisting}
You are answering a sub-question that is part of a multi-hop reasoning
chain. This sub-question is a starting step (no prior reasoning
steps).

[MAIN QUESTION]
{main_question}

[YOUR TASK]
Answer the following sub-question using the supporting paragraphs
below.

Sub-question: {root_sub_question}

[SUPPORTING PARAGRAPHS]
Some paragraphs below may be irrelevant distractors.

{paragraphs}

[INSTRUCTIONS]
- Use the paragraphs to ground your answer.
- Output the answer as a short, direct text span (no explanation, no
  reasoning trace).

[OUTPUT FORMAT]
Output only a JSON object:
{
  "answer": "<your answer here>"
}
\end{lstlisting}

\paragraph{Sampled~BN: conditional prompt.}

\begin{lstlisting}
You are answering one step of a multi-hop reasoning chain. The main
question will be answered through a sequence of intermediate
sub-questions; you are asked to answer one specific intermediate
step.

[MAIN QUESTION]
{main_question}

[CONTEXT: PREVIOUS REASONING STEPS]
The following sub-questions have already been addressed in prior
steps. For the purpose of this query, treat their answers as given
premises, even if they may or may not match what you would
independently conclude from the paragraphs.

{previous_steps}

[YOUR TASK]
Answer the following sub-question, assuming the previous steps'
answers as stated above:

Step {current_id}:
  Sub-question: {current_sub_question}

[SUPPORTING PARAGRAPHS]
Some paragraphs below may be irrelevant distractors.

{paragraphs}

[INSTRUCTIONS]
- Use the assumed answers from previous steps as given facts when
  answering.
- Use the paragraphs to ground your answer where needed.
- Output the answer as a short, direct text span (no explanation, no
  reasoning trace).
- If the assumed answers and paragraphs are jointly inconsistent,
  still produce your best answer conditional on the assumed answers;
  do not refuse to answer.

[OUTPUT FORMAT]
Output only a JSON object:
{
  "answer": "<your answer here>"
}
\end{lstlisting}

\paragraph{Dynamic fields.}

\texttt{\{question\}}: original main question.
\texttt{\{paragraphs\_block\}} / \texttt{\{paragraphs\}}: candidate paragraphs formatted as \texttt{[idx N] Title: \{title\}\textbackslash n\{text\}}, separated by blank lines.
\texttt{\{node\_list\_without\_edges\}}: gold sub-questions plus \texttt{q\_final}, without dependency edges.
\texttt{\{node\_list\}}: reasoning node ids (\texttt{q1}, \texttt{q2}, \ldots, \texttt{q\_final}).
\texttt{\{edge\_list\}}: dependency edges formatted as \texttt{qi -> qj}.
\texttt{\{previous\_steps\}}: parent-state answers already sampled in the current Sampled~BN CPT row.

\subsection{Prompt Reproducibility}
\label{app:prompt-reproducibility}

SHA256 hashes for the Stage~1 prompt templates are computed over the complete static text (excluding instance-specific fields: question, paragraphs, and node content). All hashes correspond to the prompt version used in all reported experiments.

\noindent\resizebox{\linewidth}{!}{%
  \ttfamily
  \begin{tabular}{@{}ll@{}}
  A: & 7b79d14341222c19d20bf767c12581c651120e2aec34ee9570cd97cd49b3faed \\
  B: & 05d8aabbc86bbec6dbf8d5a9d52cfc48cef5ac2e98f8f12c4c911cba3111bfcb \\
  C: & ba1a85ccc65a8d04a3f94a885c4c9c67aac77afcade5099c3859b2c65ca07a9a \\
  \end{tabular}%
}

Stage~1 uses temperature~0; Qwen3 models use \texttt{enable\_thinking=false}. Stage~2 sampling methods use temperature~0.7 with $K{=}10$ samples per call. JSON-mode output enforcement is applied to Stage~1 structured outputs where supported by the inference backend; Stage~2 methods use standard chat completions.

% ============================================================
\subsection{Parse and Schema Validity}
\label{app:reproducibility}
% ============================================================

Table~\ref{tab:reproducibility} reports parse and schema validity statistics across the archived full-dev runs (2,417 examples per model and setting).

\begin{table}[h]
\centering
\small
\resizebox{\linewidth}{!}{%
\begin{tabular}{ll rr}
\toprule
\textbf{Model} & \textbf{Setting} &
  \textbf{Parse OK} & \textbf{Schema OK} \\
\midrule
Llama-3.1-8B   & A & 1.000 & 1.000 \\
               & B & 0.998 & 0.998 \\
               & C & 0.990 & 0.990 \\
\midrule
Llama-3.1-70B  & A & 1.000 & 1.000 \\
               & B & 0.996 & 0.996 \\
               & C & 0.995 & 0.995 \\
\midrule
Qwen3-32B      & A & 0.999 & 0.999 \\
               & B & 0.994 & 0.994 \\
               & C & 0.990 & 0.990 \\
\midrule
Claude Sonnet 4.5 & A & 1.000 & 1.000 \\
                  & B & 1.000 & 1.000 \\
                  & C & 1.000 & 1.000 \\
\bottomrule
\end{tabular}}
\caption{Parse and schema validity across models and settings (2,417 examples per run).}
\label{tab:reproducibility}
\end{table}

% ============================================================
\subsection{Full Schema-Example Ablation}
\label{app:schema-full-dev}
% ============================================================

The ablation uses Setting~A on MuSiQue full development set (2,417 examples). The full Setting~A prompt is in Appendix~\ref{app:stage1-prompts}. All decoding is at temperature~0 with a 2500-token output budget and strict retry on schema incompleteness. The five variants differ only in the JSON schema example appended after the shared instruction block; all other prompt text, data, metrics, and inference code are unchanged.

\subsubsection*{Variant Definitions}

\begin{description}
\item[E1 (single-edge, current default).] Shows \texttt{q1}$\to$\texttt{q2}; \texttt{q\_final} appears in \texttt{topological\_order} but receives no explicit incoming edge in the example.
\begin{lstlisting}
{"edges": [{"from": "q1", "to": "q2",
            "confidence": 0.85,
            "justification": "q2 directly requires q1 because ..."}],
 "topological_order": ["q1", "q2", "q_final"]}
\end{lstlisting}

\item[E2 (legacy single-edge).] Shows \texttt{q1}$\to$\texttt{q3}; same structural pattern as E1 but with a non-adjacent target node.
\begin{lstlisting}
{"edges": [{"from": "q1", "to": "q3",
            "confidence": 0.85,
            "justification": "q3 directly requires q1 because ..."}],
 "topological_order": ["q1", "q3", "q_final"]}
\end{lstlisting}

\item[E3 (placeholder chain).] Uses placeholder node names (\texttt{qA}, \texttt{qB}) that cannot be confused with instance IDs, and explicitly connects to \texttt{q\_final}.
\begin{lstlisting}
{"edges": [{"from": "qA", "to": "qB",
            "confidence": 0.85,
            "justification": "qB directly requires qA because ..."},
           {"from": "qB", "to": "q_final",
            "confidence": 0.9,
            "justification": "q_final directly requires qB because ..."}],
 "topological_order": ["qA", "qB", "q_final"]}
\end{lstlisting}

\item[E4 (placeholder v-structure).] Shows a converging dependency (two independent nodes sharing a common successor), the structure hardest to elicit from chain-shaped prompts.
\begin{lstlisting}
{"edges": [{"from": "qA", "to": "qC", "confidence": 0.85,
            "justification": "qC directly requires qA because ..."},
           {"from": "qB", "to": "qC", "confidence": 0.85,
            "justification": "qC directly requires qB because ..."},
           {"from": "qC", "to": "q_final", "confidence": 0.9,
            "justification": "q_final directly requires qC because ..."}],
 "topological_order": ["qA", "qB", "qC", "q_final"]}
\end{lstlisting}

\item[E5 (format-only placeholder).] Replaces all node names with angle-bracket tokens, providing only the field-format skeleton with no topological signal.
\begin{lstlisting}
{"edges": [{"from": "<source_id>", "to": "<target_id>",
            "confidence": 0.85,
            "justification": "..."}],
 "topological_order": ["..."]}
\end{lstlisting}
\end{description}

\subsubsection*{Full Results}

Table~\ref{tab:schema-ablation-full} reports all five variants across four models.

% Full A_v2 schema-example ablation across model families.
% N = 2417 per variant per model. Setting A_v2, dependency wiring with gold nodes.
% Only the output-schema example differs across variants (E1--E5).
\begin{table*}[t]
\centering
\small
\setlength{\tabcolsep}{4pt}
\resizebox{\linewidth}{!}{%
\begin{tabular}{ll r rrrrrr}
\toprule
\textbf{Model} & \textbf{Var.} & \textbf{N} &
  \textbf{Parse OK} & \textbf{Schema OK} & \textbf{Disconn.} &
  \textbf{Edge F1} & \textbf{Exact Edge} & \textbf{Shortcut} \\
\midrule
Qwen3-8B & E1 & 2417 & 0.998 & 0.998 & 0.472 & 0.809 & 0.444 & 0.058 \\
         & E2 & 2417 & 0.998 & 0.998 & 0.260 & 0.856 & 0.611 & 0.097 \\
         & E3 & 2417 & 0.998 & 0.998 & 0.013 & 0.909 & 0.793 & 0.153 \\
         & E4 & 2417 & 0.998 & 0.998 & 0.008 & 0.902 & 0.785 & 0.169 \\
         & E5 & 2417 & 0.998 & 0.998 & 0.281 & 0.836 & 0.550 & 0.126 \\
\midrule
Qwen3-32B & E1 & 2417 & 0.999 & 0.999 & 0.003 & 0.985 & 0.961 & 0.028 \\
          & E2 & 2417 & 0.999 & 0.999 & 0.003 & 0.984 & 0.960 & 0.025 \\
          & E3 & 2417 & 0.999 & 0.999 & 0.002 & 0.987 & 0.966 & 0.024 \\
          & E4 & 2417 & 0.998 & 0.998 & 0.002 & 0.986 & 0.963 & 0.028 \\
          & E5 & 2417 & 0.999 & 0.999 & 0.001 & 0.982 & 0.956 & 0.033 \\
\midrule
Llama-3.1-8B & E1 & 2417 & 1.000 & 1.000 & 0.474 & 0.797 & 0.387 & 0.024 \\
             & E2 & 2417 & 0.999 & 0.996 & 0.094 & 0.876 & 0.684 & 0.056 \\
             & E3 & 2417 & 0.999 & 0.998 & 0.042 & 0.920 & 0.791 & 0.036 \\
             & E4 & 2417 & 1.000 & 0.995 & 0.145 & 0.888 & 0.689 & 0.028 \\
             & E5 & 2417 & 0.983 & 0.729 & 0.292 & 0.675 & 0.483 & 0.086 \\
\midrule
Llama-3.1-70B & E1 & 2417 & 1.000 & 1.000 & 0.014 & 0.899 & 0.778 & 0.153 \\
              & E2 & 2417 & 1.000 & 1.000 & 0.010 & 0.911 & 0.797 & 0.139 \\
              & E3 & 2417 & 1.000 & 1.000 & 0.003 & 0.940 & 0.870 & 0.086 \\
              & E4 & 2417 & 1.000 & 1.000 & 0.002 & 0.936 & 0.857 & 0.096 \\
              & E5 & 2417 & 1.000 & 1.000 & 0.002 & 0.897 & 0.784 & 0.163 \\
\bottomrule
\multicolumn{9}{l}{\footnotesize Variants: E1 current q1$\to$q2 single edge; E2 legacy q1$\to$q3 single edge;
  E3 placeholder chain qA$\to$qB$\to$q\_final;} \\
\multicolumn{9}{l}{\footnotesize E4 placeholder v-structure qA/qB$\to$qC$\to$q\_final;
  E5 format-only single placeholder.} \\
\end{tabular}}
\caption{Full Setting~A schema-example ablation across model families on MuSiQue full development set
  (2,417 examples per model/variant). Variants differ only in the output-schema example;
  all other prompt text, dataset, metrics, and inference code are unchanged.
  E1 is the variant used in main experiments (Table~\ref{tab:stage1-main-results}).}
\label{tab:schema-ablation-full}
\end{table*}

\section{Metric Definitions}
\label{app:metrics}

\subsection*{Node Alignment (Setting C)}

In Setting C the LLM generates its own sub-questions, so predicted nodes must be aligned to gold nodes before any downstream metric is computed.

% \paragraph{Paragraph-ID matching (primary).} Node similarity is defined as the Jaccard coefficient over supporting paragraph ID sets:
% \[
% \text{sim}_{\text{para}}(v_i, u_j) = \frac{|\text{evidence}(v_i) \cap \text{evidence}(u_j)|}{|\text{evidence}(v_i) \cup \text{evidence}(u_j)|}
% \]
% Maximum-weight bipartite matching is computed over these similarities; a pair is considered matched if $\text{sim}_{\text{para}} > 0$. This criterion is discrete and requires no threshold tuning. It aligns with the HPS formulation of \citet{ning_mc-search_2026}.

\paragraph{Semantic matching.} Nodes are aligned via sub-question text similarity using e5-large-v2 embeddings \citep{wang2022text} with cosine similarity at threshold $\tau = 0.85$. This captures cases where two nodes address the same reasoning step but cite different paragraph indices. All Setting~C results in the paper use e5-large-v2 semantic matching at $\tau{=}0.85$.

\subsection*{Dependency Wiring Metrics}

\paragraph{Edge HPS (Setting C).}
\[
\mathrm{Edge\ HPS} = \frac{|\{(u,v)\in E: \hat{m}(u)\leadsto \hat{m}(v)\}|}{|E|},
\]
where $\hat{m}(u)\in\hat{V}$ is the predicted node matched to gold node $u$, $\leadsto$ denotes directed reachability in the predicted graph, and the denominator $|E|$ is the number of gold dependency edges. Reachability (rather than direct edge existence) is used because the LLM may decompose one gold step into multiple finer steps, producing an indirect but valid dependency path. The final-answer node $v_{\mathrm{final}}$ is handled separately: if the predicted graph contains a final-answer node, it is force-aligned to the gold $v_{\mathrm{final}}$ regardless of semantic similarity; if no final-answer node is present, all gold edges incident to $v_{\mathrm{final}}$ are counted as misses.

\subsection*{Node Generation Metrics (Setting C)}

\paragraph{HPS / Step F1.} Step precision is the fraction of predicted nodes matching a gold node; Step F1 is the harmonic mean of HPS and step precision. Both metrics exclude $v_{\mathrm{final}}$ and its predicted counterpart from $V$ and $\hat{V}$: only intermediate step nodes are counted.

\subsection*{Evidence Grounding Metrics}

\paragraph{Attribution F1.} The final-answer node $v_{\mathrm{final}}$ is excluded from attribution, as its evidence spans all intermediate steps rather than a specific supporting paragraph.

\subsection*{Confidence Quality Metrics (Stage~2)}

\paragraph{Expected Calibration Error (ECE).} Predictions are binned by confidence; ECE is the weighted average of the absolute difference between mean confidence and mean accuracy within each bin.

\paragraph{Brier Score.} The mean squared error between predicted confidence and binary correctness label, measuring both calibration and sharpness.

\paragraph{AUROC.} Area under the ROC curve, measuring the model's ability to discriminate correct from incorrect answers using confidence scores as a ranking signal.

\subsection*{Format Validity and Invalid-Output Handling}

Parse rate, schema validity, acyclicity, and illegal-edge rate are computed before any structural metric. Invalid or malformed outputs are not silently dropped: each failure mode is counted in the corresponding validity metric. Outputs that fail to parse are excluded from downstream structural and calibration metrics, and the fraction excluded is reported alongside those metrics.

% ============================================================
\section{Stage 1 Supplementary Results}
\label{app:stage1-supplementary}
% ============================================================

% ============================================================
\subsection{Schema Sensitivity in Fixed-Node Edge Wiring}
\label{app:schema-sensitivity}
% ============================================================

Setting~A isolates fixed-node edge wiring, but the schema example in the prompt can still bias graph generation. Table~\ref{tab:schema-ablation-full} reports the full ablation across five variants (E1--E5) and four models; only the embedded schema example differs across variants.

E1 uses a concrete topology where \texttt{q\_final} appears in the topological order but has no explicit incoming edge. E3 uses placeholder node names and explicitly connects to \texttt{q\_final}. Small models strongly mimic E1: Qwen3-8B and Llama-3.1-8B show disconnected rates of 0.472 and 0.474 under E1, dropping to 0.013 and 0.042 under E3. E4, a v-structure placeholder, achieves similarly low disconnected rates (Qwen3-8B: 0.008; Llama-3.1-8B: 0.145). E5, a format-only placeholder, causes partial schema failure for Llama-3.1-8B (Schema OK: 0.729). Larger models are more robust: Qwen3-32B varies by less than 0.5~pp in Edge F1 across all variants, while Llama-3.1-70B still gains 9.2~pp exact-edge match under E3 relative to E1.

% ============================================================
\subsection{Thinking-On vs.\ Non-Thinking: Qwen3-32B}
\label{app:thinking}
% ============================================================

Table~\ref{tab:thinking} compares Qwen3-32B under non-thinking mode (the default used in main experiments) and thinking-on mode, both under Setting~A (2,417 examples). These are separate decoding conditions; results are not pooled.

Thinking mode yields small but consistent gains on an already strong model: Edge F1 improves by 0.7~pp, exact edge match by 1.5~pp, and the shortcut rate falls by 1.0~pp. Because non-thinking Qwen3-32B is already near-ceiling on Setting~A, the absolute gains are modest. We use non-thinking as the main-experiment default because the marginal improvement does not justify the added compute cost and latency.

\begin{table}[h]
\centering
\small
\resizebox{\linewidth}{!}{%
\begin{tabular}{lrrr}
\toprule
\textbf{Decoding} & \textbf{Edge F1} & \textbf{Exact Edge} & \textbf{Shortcut} \\
\midrule
Non-thinking (main) & 0.985 & 0.961 & 0.028 \\
Thinking-on         & 0.992 & 0.976 & 0.018 \\
\bottomrule
\end{tabular}}
\caption{Qwen3-32B thinking-on vs.\ non-thinking under Setting~A (2,417 examples).
  Thinking mode yields a further 0.7~pp Edge F1 gain and 1~pp shortcut reduction.}
\label{tab:thinking}
\end{table}

\subsection{Strict Direct-Edge Evaluation}
\label{app:strict-edge}

The main Edge HPS metric counts a gold dependency as recovered when the
aligned predicted nodes are connected by a directed path, allowing for
finer-grained intermediate decompositions. To test whether this
reachability-based criterion is overly permissive, we recompute Edge HPS
using a stricter criterion that requires a direct predicted edge between
the aligned endpoints. 

\begin{table}[t]
\centering
\small
\setlength{\tabcolsep}{5pt}
\begin{tabular}{lcc}
\toprule
\textbf{Model} & \textbf{Edge HPS} & \textbf{Strict Edge HPS} \\
\midrule
Llama-3.1-70B     & 0.612 & 0.597 \\
Qwen3-32B         & 0.705 & 0.672 \\
Claude Sonnet 4.5 & 0.890 & 0.841 \\
\bottomrule
\end{tabular}
\caption{Setting C Edge HPS under the main reachability-based criterion
and a strict direct-edge criterion on the MuSiQue full development set.}
\label{tab:strict-edge}
\end{table}

As shown in Table~\ref{tab:strict-edge}, the stricter criterion lowers
absolute scores by 1.5--4.9 points but preserves the same model ordering.
This indicates that the relative structural comparison is not driven by
the reachability-based definition of Edge HPS.

\subsection{Node-Alignment Sensitivity}
\label{app:alignment-sensitivity}

Setting C requires semantic alignment between generated and gold
subquestions before computing HPS-based metrics. The main experiments use
e5-large-v2 embeddings with cosine-similarity threshold $\tau=0.85$.
To assess sensitivity to this choice, we recompute HPS over the full
MuSiQue development set using
$\tau \in \{0.75, 0.85, 0.90, 0.95\}$ and three embedding backbones.
Table~\ref{tab:alignment-sensitivity} reports mean HPS across the nine
models in Table~\ref{tab:stage1-main-results}.

\begin{table}[t]
\centering
\small
\setlength{\tabcolsep}{5pt}
\resizebox{\linewidth}{!}{
\begin{tabular}{lcccc}
\toprule
\textbf{Backbone}
& $\boldsymbol{\tau=0.75}$
& $\boldsymbol{\tau=0.85}$
& $\boldsymbol{\tau=0.90}$
& $\boldsymbol{\tau=0.95}$ \\
\midrule
e5-large-v2
    & 0.913 & 0.836 & 0.714 & 0.504 \\
MiniLM-L6-v2
    & 0.703 & 0.523 & 0.378 & 0.176 \\
BGE-large-v1.5
    & 0.787 & 0.621 & 0.474 & 0.230 \\
\bottomrule
\end{tabular}}
\caption{
Mean Setting C HPS across the nine main-table models under different
semantic-alignment thresholds and embedding backbones.
}
\label{tab:alignment-sensitivity}
\end{table}

As shown in Table~\ref{tab:alignment-sensitivity}, absolute HPS values
depend substantially on the alignment protocol: lower thresholds produce
more permissive matches, while higher thresholds reject more paraphrastic
matches. However, relative model rankings remain stable. At $\tau=0.85$,
the pairwise Spearman correlations of model rankings across the three
backbones are 0.900, 0.967, and 0.950. Thus, Setting C HPS should be
interpreted as a reference-alignment diagnostic under the stated protocol,
while the relative comparisons across models are substantially more stable
than the absolute scores.

% ============================================================
\subsection{Stage 1 Error Diagnostics}
\label{app:stage1-error-diagnostics}
% ============================================================

\begin{figure}[h]
    \centering
    \includegraphics[width=\linewidth]{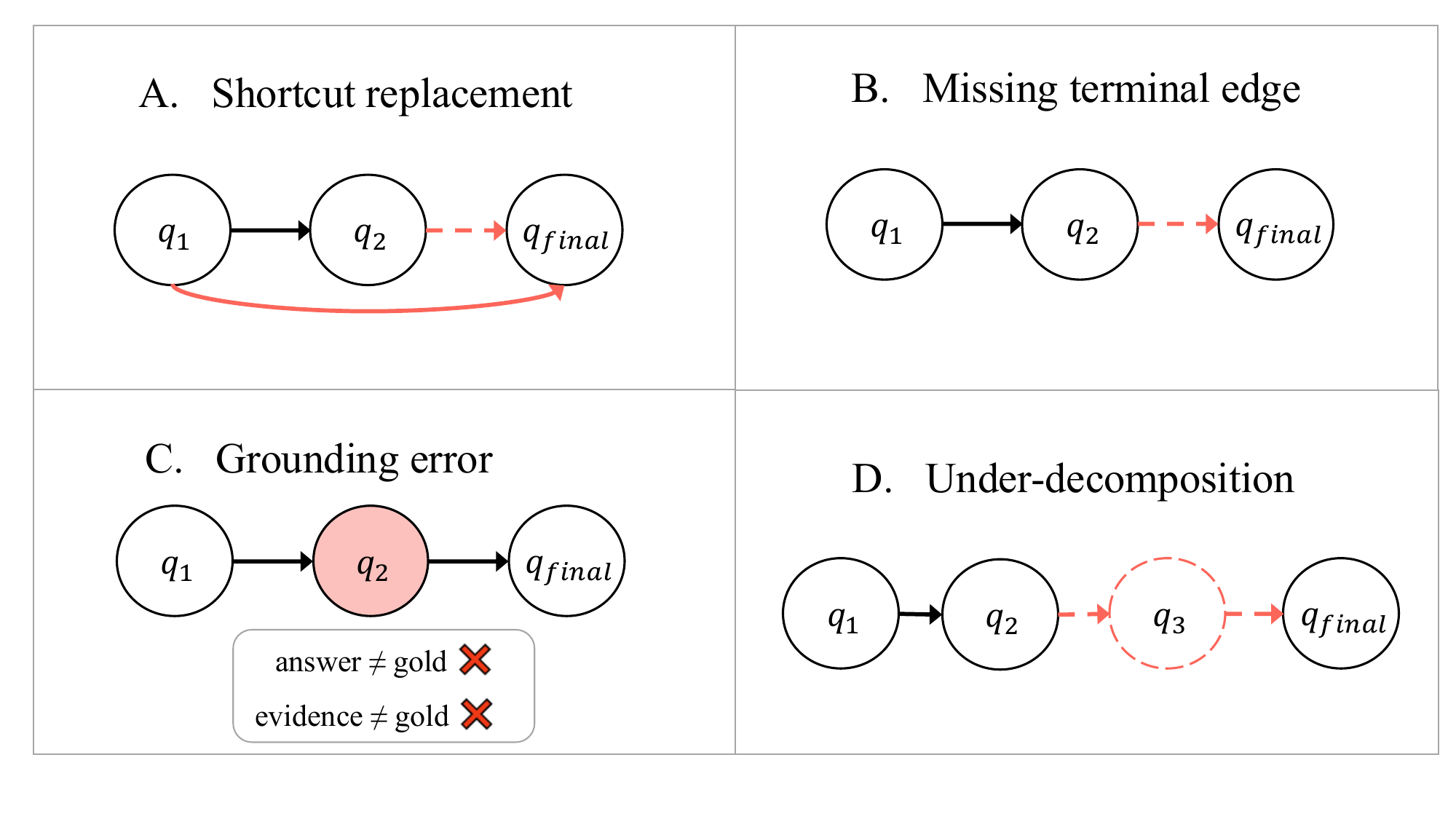}
    \caption{Schematic error modes across diagnostic settings. Red solid edges indicate incorrect dependencies, red dashed edges or nodes indicate missing structure, and shaded nodes indicate incorrect local payloads.}
    \label{fig:stage1-error-modes}
\end{figure}

This appendix defines the automatic diagnostic tags used to summarize Stage~1 failures. The tags are operational labels for Stage~1 outputs, not a complete taxonomy of all possible LLM reasoning failures. Figure~\ref{fig:stage1-error-modes} illustrates the three main failure modes schematically.

\begin{table*}[h]
\centering
\small
\setlength{\tabcolsep}{4pt}
\begin{tabular}{p{0.24\textwidth}p{0.70\textwidth}}
\toprule
\textbf{Tag} & \textbf{Operational Detection Criterion} \\
\midrule
fully\_aligned & Predicted nodes, edges, answers, and evidence all match the gold graph under the setting-specific evaluator. \\
shortcut\_replacement & A gold dependency is missing and a direct or transitive shortcut edge connects an earlier prerequisite to a later node, often \texttt{q\_final}. \\
spurious\_final\_shortcut & A non-parent node has an extra direct edge into \texttt{q\_final}. \\
spurious\_internal\_shortcut & A non-parent node has an extra shortcut edge into an internal node. \\
mixed\_shortcut & Final and internal shortcut behavior appear in the same fixed-node output. \\
missing\_terminal\_edge & A gold dependency into \texttt{q\_final} is omitted. \\
missing\_gold\_edge & A required gold dependency is absent and is not captured by a more specific missing-terminal or shortcut tag. \\
other\_edge\_error & The predicted edge set differs from gold but does not satisfy the shortcut or missing-edge criteria above. \\
answer\_and\_evidence & A fixed gold sub-question has both an incorrect generated answer and incorrect evidence attribution. \\
evidence\_only & The answer is correct or matched, but the evidence attribution is wrong. \\
answer\_generation\_only & The supporting evidence is correct or matched, but the generated answer is wrong. \\
\makecell[l]{intermediate\_answer\_only\\\_recovered\_final} & One or more intermediate answers are wrong, but the final answer is recovered. \\
edge\_reasoning\_only & Local node content is correct or matched, but the dependency structure is wrong. \\
broad\_mixed & Multiple answer, evidence attribution, and edge failures occur in the same output without one dominant local cause. \\
answer\_and\_edge & Generated answers and dependency edges are wrong while evidence is not the only failure. \\
edge\_and\_evidence & Dependency edges and evidence attribution are wrong while generated answers are not the only failure. \\
subquestion\_drift & The fixed sub-question is paraphrased or followed in a way that changes the intended local reasoning step. \\
final\_synthesis\_only & Intermediate node content is acceptable, but the final synthesized answer is wrong. \\
final\_plus\_evidence & Final answer synthesis and evidence attribution both fail. \\
final\_plus\_edge & Final answer synthesis and edge recovery both fail. \\
final\_mixed & Final answer synthesis fails together with multiple other local errors. \\
under\_decompose & The predicted graph contains fewer reasoning steps than the gold decomposition. \\
over\_decompose & The predicted graph contains extra reasoning steps beyond the gold decomposition. \\
structure\_gap & Predicted steps are plausible or align to gold steps, but the dependency structure among them is wrong. \\
evidence\_gap & Predicted steps and structure align, but evidence attribution fails. \\
extra\_benign & Extra predicted steps are present without degrading the matched structural or evidence metrics. \\
minor\_mixed & A residual mixed category for small combined C-setting failures that do not dominate the output. \\
parse\_failure & The model output cannot be parsed into the expected JSON object. \\
schema\_failure & The parsed output violates the required schema, for example by omitting required fields or using invalid node identifiers. \\
\bottomrule
\end{tabular}
\caption{Stage~1 diagnostic failure taxonomy. Tags are automatic labels over the evaluated graph outputs, intended to localize failure modes within each diagnostic setting.}
\label{tab:stage1-error-taxonomy}
\end{table*}

\paragraph{Full-dev count evidence.}
Tables~\ref{tab:stage1-error-counts-a}--\ref{tab:stage1-error-counts-c} report full-dev error counts by model for Settings~A--C. Counts cover the full development set and use the same prompt schemas as all reported experiments.

% Full-dev Stage 1 error counts from reports/stage1_full_dev_comparison_2026-05-21/stage1_full_dev_error_analysis_report.md.

\begin{table*}[p]
\centering
\scriptsize
\setlength{\tabcolsep}{3pt}
\resizebox{\textwidth}{!}{%
\begin{tabular}{lrrrrrrrrr}
\toprule
\textbf{Model} & \textbf{Full} & \textbf{Shortcut repl.} & \textbf{Missing term.} & \textbf{Missing gold} & \textbf{Other edge} & \textbf{Final shortcut} & \textbf{Internal shortcut} & \textbf{Mixed shortcut} & \textbf{Parse fail.} \\
\midrule
Claude Sonnet~4.5 & 2382 & 32 & 0 & 2 & 0 & 1 & 0 & 0 & 0 \\
DeepSeek V4 Flash & 2372 & 45 & 0 & 0 & 0 & 0 & 0 & 0 & 0 \\
GPT-5 Chat & 2337 & 67 & 0 & 0 & 2 & 10 & 0 & 1 & 0 \\
Llama-3.1-70B & 1880 & 465 & 3 & 48 & 8 & 9 & 4 & 0 & 0 \\
Llama-3.1-8B & 937 & 89 & 1105 & 23 & 245 & 5 & 0 & 13 & 0 \\
Qwen3-8B & 1071 & 173 & 1123 & 11 & 20 & 7 & 0 & 6 & 6 \\
Qwen3-14B & 2203 & 172 & 9 & 20 & 10 & 0 & 0 & 0 & 3 \\
Qwen3-32B & 2319 & 78 & 0 & 13 & 4 & 0 & 0 & 0 & 3 \\
Qwen3-32B-Reasoning & 2342 & 44 & 0 & 12 & 1 & 0 & 0 & 0 & 18 \\
\bottomrule
\end{tabular}}
\caption{Full-dev Setting~A error counts by model (2,417 examples). Full denotes \texttt{fully\_aligned}. Rates are reported in the source analysis; this table gives counts to avoid rounding ambiguity.}
\label{tab:stage1-error-counts-a}
\end{table*}

\begin{table*}[p]
\centering
\scriptsize
\setlength{\tabcolsep}{2pt}
\resizebox{\textwidth}{!}{%
\begin{tabular}{lrrrrrrrrrrrrrrrr}
\toprule
\textbf{Model} & \textbf{Full} & \textbf{Ans+Ev} & \textbf{Ev only} & \textbf{Ans only} & \textbf{Interm. ans} & \textbf{Edge only} & \textbf{Broad} & \textbf{Ans+Edge} & \textbf{Edge+Ev} & \textbf{Subq drift} & \textbf{Final synth} & \textbf{Final+Ev} & \textbf{Final+Edge} & \textbf{Final mixed} & \textbf{Parse} & \textbf{Schema} \\
\midrule
Claude Sonnet~4.5 & 1195 & 449 & 390 & 145 & 142 & 21 & 9 & 8 & 5 & 36 & 7 & 6 & 4 & 0 & 0 & 0 \\
DeepSeek V4 Flash & 1018 & 591 & 371 & 170 & 168 & 24 & 21 & 25 & 9 & 4 & 7 & 4 & 3 & 0 & 2 & 0 \\
GPT-5 Chat & 839 & 584 & 498 & 177 & 172 & 28 & 29 & 28 & 12 & 0 & 23 & 19 & 6 & 2 & 0 & 0 \\
Llama-3.1-70B & 756 & 572 & 290 & 160 & 108 & 152 & 157 & 101 & 38 & 54 & 3 & 6 & 10 & 1 & 9 & 0 \\
Llama-3.1-8B & 54 & 630 & 60 & 37 & 58 & 210 & 847 & 175 & 144 & 168 & 6 & 12 & 9 & 2 & 5 & 0 \\
Qwen3-8B & 543 & 983 & 317 & 183 & 130 & 15 & 111 & 42 & 9 & 51 & 7 & 4 & 2 & 1 & 16 & 3 \\
Qwen3-14B & 640 & 822 & 340 & 196 & 98 & 48 & 83 & 53 & 22 & 83 & 11 & 2 & 3 & 0 & 13 & 3 \\
Qwen3-32B & 793 & 730 & 300 & 213 & 133 & 25 & 68 & 30 & 13 & 80 & 12 & 3 & 3 & 0 & 14 & 0 \\
Qwen3-32B-Reasoning & 887 & 626 & 351 & 163 & 123 & 27 & 87 & 43 & 16 & 27 & 8 & 4 & 1 & 0 & 52 & 2 \\
\bottomrule
\end{tabular}}
\caption{Full-dev Setting~B error counts by model (2,417 examples). Full denotes \texttt{fully\_aligned}; Ans+Ev denotes \texttt{answer\_and\_evidence}; Interm. ans denotes \texttt{intermediate\_answer\_only\_recovered\_final}.}
\label{tab:stage1-error-counts-b}
\end{table*}

\begin{table*}[p]
\centering
\scriptsize
\setlength{\tabcolsep}{3pt}
\resizebox{\textwidth}{!}{%
\begin{tabular}{lrrrrrrrrr}
\toprule
\textbf{Model} & \textbf{Full} & \textbf{Under} & \textbf{Structure gap} & \textbf{Evidence gap} & \textbf{Over} & \textbf{Extra benign} & \textbf{Minor mixed} & \textbf{Parse fail.} & \textbf{Schema fail.} \\
\midrule
Claude Sonnet~4.5 & 1172 & 99 & 228 & 452 & 221 & 226 & 18 & 1 & 0 \\
DeepSeek V4 Flash & 900 & 506 & 393 & 306 & 164 & 91 & 19 & 9 & 29 \\
GPT-5 Chat & 873 & 153 & 423 & 421 & 291 & 221 & 21 & 0 & 14 \\
Llama-3.1-70B & 701 & 838 & 454 & 278 & 72 & 30 & 33 & 11 & 0 \\
Llama-3.1-8B & 134 & 639 & 917 & 203 & 408 & 62 & 31 & 23 & 0 \\
Qwen3-8B & 602 & 932 & 424 & 354 & 30 & 11 & 39 & 22 & 3 \\
Qwen3-14B & 647 & 815 & 515 & 321 & 64 & 19 & 21 & 11 & 4 \\
Qwen3-32B & 810 & 373 & 471 & 396 & 206 & 113 & 25 & 23 & 0 \\
Qwen3-32B-Reasoning & 914 & 501 & 305 & 457 & 106 & 49 & 16 & 63 & 6 \\
\bottomrule
\end{tabular}}
\caption{Full-dev Setting~C error counts by model (2,417 examples). Full denotes \texttt{fully\_aligned}; Under and Over denote \texttt{under\_decompose} and \texttt{over\_decompose}.}
\label{tab:stage1-error-counts-c}
\end{table*}

% ============================================================
\subsection{Within-Hop Topology Breakdown}
\label{app:within-hop-topology}
% ============================================================

Topology buckets are defined as follows: \textit{simple\_tree} requires \texttt{max\_indegree}$<$2, \texttt{max\_depth}$<$3, and \texttt{num\_nodes}$<$5; \textit{v\_structure} has exactly one merge node with parallel parents; remaining graphs are \textit{complex\_tree}. MuSiQue hop count and topology are not fully crossed: all 2-hop examples are simple\_tree, 3-hop examples contain simple\_tree and v\_structure, and 4-hop examples contain v\_structure and complex\_tree. We therefore compare topology only within hop strata where multiple topology buckets are available.

\begin{table*}[htbp]
\centering
\small
\setlength{\tabcolsep}{4pt}
\resizebox{0.9\textwidth}{!}{
\begin{tabular}{lllrrrrr}
\toprule
\textbf{Model} & \textbf{Hop} & \textbf{Topology} & \textbf{N} & \textbf{HPS} & \textbf{Edge HPS} & \textbf{Evidence HPS} & $\boldsymbol{\Delta}$\textbf{Step} \\
\midrule
Llama-3.1-70B & 3 & simple\_tree & 568 & 0.715 & 0.517 & 0.667 & -0.626 \\
Llama-3.1-70B & 3 & v\_structure & 192 & 0.717 & 0.370 & 0.568 & -0.609 \\
Llama-3.1-70B & 4 & v\_structure & 159 & 0.642 & 0.326 & 0.566 & -0.861 \\
Llama-3.1-70B & 4 & complex\_tree & 246 & 0.556 & 0.349 & 0.535 & -1.193 \\
\midrule
Qwen3-32B & 3 & simple\_tree & 568 & 0.842 & 0.692 & 0.742 & -0.073 \\
Qwen3-32B & 3 & v\_structure & 192 & 0.842 & 0.477 & 0.696 & 0.052 \\
Qwen3-32B & 4 & v\_structure & 159 & 0.733 & 0.426 & 0.610 & -0.302 \\
Qwen3-32B & 4 & complex\_tree & 246 & 0.714 & 0.511 & 0.686 & -0.459 \\
\midrule
Claude Sonnet 4.5 & 3 & simple\_tree & 568 & 0.957 & 0.888 & 0.863 & 0.192 \\
Claude Sonnet 4.5 & 3 & v\_structure & 192 & 0.977 & 0.863 & 0.821 & 0.292 \\
Claude Sonnet 4.5 & 4 & v\_structure & 159 & 0.932 & 0.803 & 0.808 & 0.516 \\
Claude Sonnet 4.5 & 4 & complex\_tree & 246 & 0.907 & 0.821 & 0.783 & -0.061 \\
\bottomrule
\end{tabular}
}
\caption{Within-hop Setting~C graph generation performance by topology on MuSiQue. MuSiQue does not contain all hop--topology combinations: 2-hop examples are all simple\_tree, 3-hop examples compare simple\_tree and v\_structure, and 4-hop examples compare v\_structure and complex\_tree. $\Delta$Step is predicted minus gold node count; negative values indicate under-decomposition.}
\label{tab:within-hop-topology}
\end{table*}

Among 3-hop questions, \textit{v\_structure} leaves HPS nearly unchanged relative to \textit{simple\_tree} (Llama-3.1-70B: 0.715 vs.\ 0.717; Qwen3-32B: 0.842 vs.\ 0.842; Claude Sonnet~4.5: 0.957 vs.\ 0.977) but lowers Edge HPS substantially for all three models (Llama: 0.517$\to$0.370; Qwen: 0.692$\to$0.477; Claude: 0.888$\to$0.863). Merge structures therefore primarily challenge dependency recovery rather than step coverage. Among 4-hop questions, \textit{complex\_tree} lowers both HPS and $\Delta$Step relative to \textit{v\_structure} for Llama-3.1-70B ($\Delta$Step $-1.193$ vs.\ $-0.861$) and Qwen3-32B ($-0.459$ vs.\ $-0.302$), while Edge HPS does not decline uniformly: for Qwen3-32B, \textit{complex\_tree} Edge HPS (0.511) is higher than \textit{v\_structure} (0.426). Deeper or multi-merge structures therefore induce under-decomposition errors rather than edge-wiring failures.

% ============================================================
\subsection{2WikiMultiHopQA Stage~1 Diagnostics}
\label{app:twowiki-stage1}
% ============================================================

Table~\ref{tab:twowiki-stage1-all} reports Stage~1 A/B/C diagnostics on the 498-example stratified 2WikiMultiHopQA development subset for five models. Because 2WikiMultiHopQA does not supply MuSiQue-style explicit reasoning nodes, gold nodes are constructed from evidence triples and evidence identifiers correspond to context-paragraph indices rather than sentence-level supporting facts. Ans~F1 in Setting~B is token-level F1 over intermediate sub-answers under the fixed-decomposition interface. Parse rates are 99.4\% or above for all runs; no run is excluded from metrics on parse-rate grounds. Metric names follow Table~\ref{tab:stage1-main-results}.

\begin{table*}[t]
\centering
\small
\setlength{\tabcolsep}{4pt}
\resizebox{\textwidth}{!}{%
\begin{tabular}{llr rr rrr rrrr}
\toprule
& & &
  \multicolumn{2}{c}{\textbf{Setting A}} &
  \multicolumn{3}{c}{\textbf{Setting B}} &
  \multicolumn{4}{c}{\textbf{Setting C}} \\
\cmidrule(lr){4-5}\cmidrule(lr){6-8}\cmidrule(lr){9-12}
\textbf{Model} & \textbf{Family} & \textbf{Params} &
  \textbf{Edge F1} & \textbf{Exact Edge} &
  \textbf{Edge F1} & \textbf{Ans F1} & \textbf{Attr F1} &
  \textbf{HPS} & \textbf{Step F1} & \textbf{Edge HPS} & \textbf{Attr F1} \\
\midrule
Llama-3.1-8B  & Llama & 8B  & 0.803 & 0.402 & 0.863 & 0.665 & 0.643 & 0.818 & 0.727 & 0.361 & 0.769 \\
Qwen3-8B      & Qwen  & 8B  & 0.923 & 0.774 & 0.863 & 0.807 & 0.724 & 0.866 & 0.903 & 0.722 & 0.901 \\
Qwen3-14B     & Qwen  & 14B & 0.972 & 0.952 & 0.799 & 0.723 & 0.756 & 0.956 & 0.960 & 0.784 & 0.920 \\
Qwen3-32B     & Qwen  & 32B & \textbf{0.988} & \textbf{0.976} & \textbf{0.932} & 0.691 & \textbf{0.765} & \textbf{0.981} & \textbf{0.967} & \textbf{0.923} & \textbf{0.939} \\
Llama-3.1-70B & Llama & 70B & 0.823 & 0.573 & 0.748 & \textbf{0.859} & 0.707 & 0.848 & 0.862 & 0.641 & 0.797 \\
\bottomrule
\end{tabular}}
\caption{2WikiMultiHopQA Stage~1 diagnostics on 498 stratified development examples, five models, all three settings. Bold marks the best value per column. Ans~F1 in Setting~B is token-level F1 over intermediate sub-answers under the fixed-decomposition interface. Metric names follow the main Stage~1 table (Table~\ref{tab:stage1-main-results}).}
\label{tab:twowiki-stage1-all}
\end{table*}

Setting~C Edge~HPS stratified by 2Wiki question type (Llama-3.1-70B) shows that question-type distribution shapes failure profiles independently of hop count (Table~\ref{tab:twowiki-stage1}). Bridge\_comparison questions, requiring 4-step parallel chain merging, produce the lowest Edge~HPS (0.308), while comparison questions with two independent fact retrievals reach near-ceiling (0.976). 2Wiki also surfaces reversed-edge errors in kinship-chain inference, a failure mode not prominent in MuSiQue, indicating that the dominant error type is governed by question templates and annotation scheme rather than hop depth alone.

% Table: 2WikiMultiHopQA Stage 1 Setting C results by official question type.
% n=498, Llama-3.1-70B, Version C end-to-end induction.
% Edge HPS reported at threshold 0.85.
% Inference and compositional are both chain-shaped; Edge HPS value is for both types combined.
% bridge_comparison requires 4-step parallel chain merging.
% comparison involves two independent chains merging at a comparison step.
\begin{table}[t]
\centering
\small
\setlength{\tabcolsep}{5pt}
\begin{tabular}{l r r}
\toprule
\textbf{Question Type} & \textbf{N} & \textbf{Edge HPS}$\uparrow$ \\
\midrule
inference$^\dagger$       & 80  & 0.634 \\
compositional$^\dagger$   & 80  & 0.634 \\
comparison       & 170 & \textbf{0.976} \\
bridge\_comparison & 168 & 0.308 \\
\bottomrule
\end{tabular}
\caption{2WikiMultiHopQA Stage~1 Setting~C Edge~HPS by official question type (498 examples, Llama-3.1-70B). Bold marks the highest. $^\dagger$Inference and compositional are both chain-shaped types; the Edge~HPS value is reported for both combined. Comparison questions involve two independent fact retrievals merged at a comparison step and reach near-ceiling. Bridge\_comparison questions require 4-step parallel chain merging and produce the lowest Edge~HPS.}
\label{tab:twowiki-stage1}
\end{table}

% ============================================================
\section{Stage 2 Supplementary Results}
\label{app:stage2-supplementary}
% ============================================================

% ============================================================
\subsection{Stage~2 Call Budget}
\label{app:call-budget}
% ============================================================

Table~\ref{tab:call-budget} reports API call counts for the full MuSiQue development set (2,417 examples). One API call is defined as one \texttt{chat.completions.create} request. Sampling-based baselines (Direct-SC, CoT-SC, Decomp-SC, Graph-CoT-SC, vPGM-Style) use batch sampling with $K{=}10$ completions per request, so each example costs one API call and yields ten completions. CoT-Conf uses a single completion per example. Sampled~BN is variable-cost: calls per example equal the number of root nodes plus the number of CPT parent-state entries in the gold graph, with each call again using $K{=}10$ samples.

\begin{table}[h]
\centering
\small
\setlength{\tabcolsep}{4pt}
\resizebox{\linewidth}{!}{
\begin{tabular}{lrrr}
\toprule
\textbf{Method} & \textbf{Calls / example} & \textbf{Completions / example} & \textbf{Full-dev calls} \\
\midrule
Direct-SC        & 1      & 10     & 2{,}417 \\
CoT-SC           & 1      & 10     & 2{,}417 \\
Decomp-SC        & 1      & 10     & 2{,}417 \\
Graph-CoT-SC     & 1      & 10     & 2{,}417 \\
vPGM-Style       & 1      & 10     & 2{,}417 \\
CoT-Conf         & 1      & 1      & 2{,}417 \\
Sampled BN (Qwen3-32B)     & 3.57 avg & 35.7 avg & 8{,}624 \\
Sampled BN (Llama-3.1-70B) & 4.10 avg & 41.0 avg & 9{,}906 \\
\bottomrule
\end{tabular}}
\caption{API call budget for Stage~2 methods on MuSiQue full development set (2,417 examples). SC-family runners use batch sampling ($K{=}10$ per request); Sampled~BN cost scales with graph structure.}
\label{tab:call-budget}
\end{table}

\paragraph{Sampled BN call distribution.}
Because call count depends on graph topology, Sampled~BN cost is right-skewed. Table~\ref{tab:bn-call-dist} shows the distribution across examples.

\begin{table}[h]
\centering
\small
\begin{tabular}{lrrrr}
\toprule
\textbf{Model} & \textbf{Mean} & \textbf{Median} & \textbf{Max} \\
\midrule
Qwen3-32B      & 3.568 & 3  & 71 \\
Llama-3.1-70B  & 4.098 & 3  & 54 \\
\bottomrule
\end{tabular}
\caption{Sampled~BN API calls per example on MuSiQue full dev.}
\label{tab:bn-call-dist}
\end{table}

\paragraph{Sampled BN cost by hop count.}
Call count increases with hop depth because deeper graphs have more CPT entries. Table~\ref{tab:bn-call-hop} reports mean calls per example stratified by hop count.

\begin{table}[h]
\centering
\small
\begin{tabular}{lrrr}
\toprule
\textbf{Model} & \textbf{2-hop} & \textbf{3-hop} & \textbf{4-hop} \\
\midrule
Qwen3-32B      & 2.280 & 3.974 & 6.788 \\
Llama-3.1-70B  & 2.568 & 4.546 & 7.990 \\
\bottomrule
\end{tabular}
\caption{Mean Sampled~BN API calls per example by hop count.}
\label{tab:bn-call-hop}
\end{table}

% ============================================================
\subsection{vPGM-Style Baseline Details}
\label{app:vpgm}
% ============================================================

The vPGM-Style baseline adapts the verbalized probabilistic graphical model framework of \citet{huang_bayesagent_2026} to open-ended multi-hop QA.

\paragraph{Latent variables and dependency structure.}
Three latent variables decompose the uncertainty sources in multi-hop QA:
\begin{itemize}
  \item $Z_E$ (Evidence Coverage): whether the candidate paragraphs contain sufficient evidence for all required reasoning steps.
  \item $Z_D$ (Distractor Control): whether irrelevant or misleading distractor paragraphs can be ignored.
  \item $Z_R$ (Reasoning Validity): whether the evidence can be connected into a valid multi-hop reasoning chain without logical gaps.
\end{itemize}
These form a conditional chain:
$P(Z_E \mid X)$,\quad $P(Z_D \mid Z_E, X)$,\quad $P(Z_R \mid Z_E, Z_D, X)$,\quad $P(Y \mid Z_E, Z_D, Z_R, X)$,
where $X$ denotes the question with candidate paragraphs and $Y$ denotes final-answer correctness. No gold graph structure is provided to the model.

\paragraph{Inference.}
Each call prompts the model to estimate the conditional probabilities above and produce a short final answer $\hat{y}$, all in a single JSON-structured completion. We draw $K{=}10$ completions per question at temperature~0.7, matching the SC baselines.

\paragraph{Confidence computation.}
Final-answer selection follows self-consistency clustering: the ten sampled $\hat{y}$ values are normalized and clustered, and the largest cluster's representative is the predicted answer $\hat{a} = \arg\max_c \lvert \mathcal{C}_c \rvert$. Confidence is then the mean verbalized posterior within the selected cluster:
\[
\mathrm{conf}(\hat{a})
= \frac{1}{\lvert \mathcal{C}_{\hat{a}} \rvert}
\sum_{k \in \mathcal{C}_{\hat{a}}} p_k(Y{=}\hat{a} \mid Z_E^{(k)}, Z_D^{(k)}, Z_R^{(k)}, X),
\]
where $\mathcal{C}_{\hat{a}}$ is the set of completions in the selected cluster. The individual latent probabilities $P(Z_E)$, $P(Z_D)$, $P(Z_R)$ are not combined into the confidence signal.

\paragraph{Adaptation from original vPGM.}
Our adaptation follows vPGM's latent-variable discovery and verbalized prediction-under-uncertainty framework, but adapts the final prediction step to open-ended QA through answer clustering rather than multiple-choice option scoring.

% ============================================================
\subsection{Selective Prediction Curves}
\label{app:selective}
% ============================================================

\begin{figure}[h]
\centering
\includegraphics[width=\linewidth]{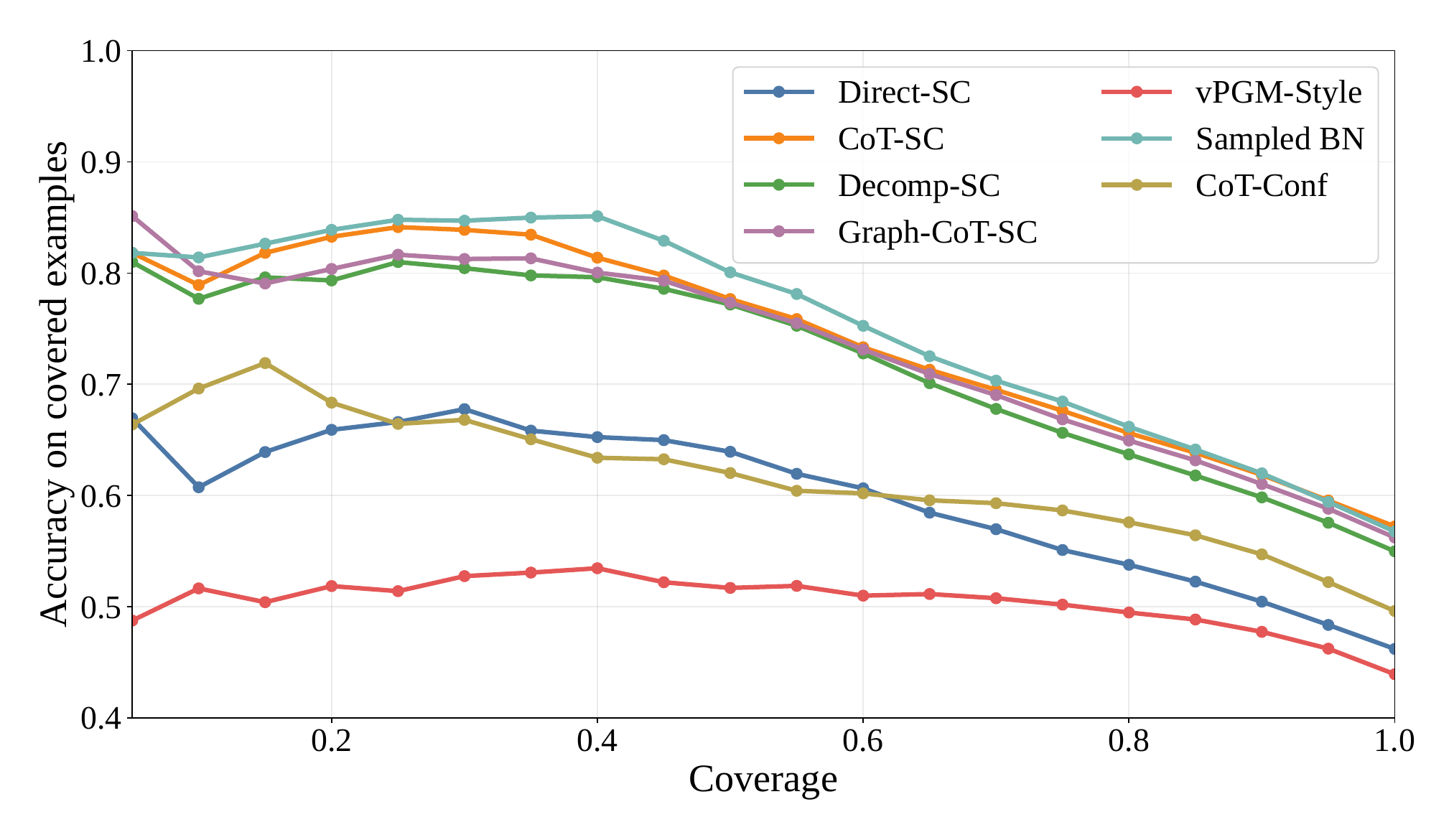}
\caption{Selective prediction curves for Stage~2 methods.}
\label{fig:stage2-selective-prediction}
\end{figure}

In selective prediction, the system answers only when its confidence exceeds a threshold and abstains otherwise. Figure~\ref{fig:stage2-selective-prediction} shows accuracy-vs-coverage curves for all Stage~2 methods. The AUROC gap between Sampled~BN and CoT-SC concentrates at intermediate coverage: at 30\% coverage the two methods are nearly tied (0.847 vs.\ 0.839), but from 30\% to 50\% coverage the gap widens to 2.4~pp. vPGM-Style stays near 0.52 accuracy across all coverage levels, consistent with its compressed confidence distribution. The two methods also produce complementary errors: Sampled~BN and CoT-SC disagree on 21.6\% of examples with a near-symmetric correctness split ($\kappa{=}0.560$; Table~\ref{tab:stage2-contingency}).

% ============================================================
\subsection{Stage 2 Per-Node Positional Breakdown}
\label{app:stage2-per-node-full}
% ============================================================

The main text (\S\ref{sec:stage2-per-node}) reports aggregate per-node diagnostics. The breakdown by chain position shows that ranking power remains stable across depth: AUROC is 0.729 at q1, 0.740 at q2, and 0.758 at q3. Absolute calibration degrades at deeper positions (q3 ECE is the worst), so per-node uncertainty should be interpreted as a reliable diagnostic ranking signal rather than a calibrated absolute probability at deeper chain positions.

% ============================================================
\subsection{Sampled BN and CoT-SC Disagreement Analysis}
\label{app:stage2-contingency}
% ============================================================

Table~\ref{tab:stage2-contingency} shows the pairwise correctness contingency between Sampled~BN and CoT-SC. On only-BN-correct examples, median Sampled~BN confidence is 0.840 versus CoT-SC's 0.600; on only-CoT-SC-correct examples, the medians reverse to 0.600 and 0.700.

\begin{table}[h]
\centering
\small
\setlength{\tabcolsep}{3pt}
\begin{tabular}{lrr}
\toprule
 & \makecell{\textbf{CoT-SC}\\\textbf{Correct}} & \makecell{\textbf{CoT-SC}\\\textbf{Wrong}} \\
\midrule
\textbf{Sampled BN Correct} & 1117 (46.2\%) & 255 (10.5\%) \\
\textbf{Sampled BN Wrong}   & 266 (11.0\%)  & 779 (32.2\%) \\
\bottomrule
\end{tabular}
\caption{Pairwise correctness contingency between Sampled~BN and CoT-SC on MuSiQue full development set (2,417 examples; $\kappa{=}0.560$). The near-symmetric one-method-only split (10.5\% BN-only correct vs.\ 11.0\% CoT-SC-only correct) indicates complementary rather than redundant error patterns.}
\label{tab:stage2-contingency}
\end{table}

\subsection{Sensitivity to CPT Smoothing}
\label{app:smoothing}

Sampled BN estimates each conditional probability table (CPT) row from
$K=10$ samples, which can yield sparse empirical counts and overconfident
probabilities. To assess this effect, we apply additive smoothing to the
cached samples from the main Llama-3.1-70B MuSiQue gold-graph experiment.
For each non-root node, we estimate

\begin{equation}
\hat{p}_{\alpha}(a_i \mid a_{\mathrm{pa}(i)})
=
\frac{
c(a_i,a_{\mathrm{pa}(i)})+\alpha
}{
\sum_{a' \in S_i} c(a',a_{\mathrm{pa}(i)})
+\alpha |S_i|
},
\end{equation}

where $S_i$ is the union of normalized answer clusters observed for node
$i$ across its CPT rows. The original Sampled BN corresponds to
$\alpha=0$.

\begin{table}[t]
\centering
\small
\setlength{\tabcolsep}{4pt}
\resizebox{\linewidth}{!}{
\begin{tabular}{c c c c c c c}
\toprule
$\alpha$
& EM
& F1
& AUROC
& ECE$\downarrow$
& Brier$\downarrow$
& High-conf.\ wrong$\downarrow$ \\
\midrule
0.0 & 0.568 & 0.691 & 0.795 & 0.206 & 0.225 & 11.0\% \\
0.1 & 0.568 & 0.690 & 0.795 & 0.194 & 0.221 & 8.0\% \\
0.5 & 0.569 & 0.692 & 0.790 & 0.168 & 0.212 & 7.3\% \\
1.0 & 0.568 & 0.690 & 0.790 & 0.147 & 0.207 & 7.0\% \\
\bottomrule
\end{tabular}}
\caption{
Sensitivity of Sampled BN to additive CPT smoothing on the
Llama-3.1-70B MuSiQue gold-graph setting.
High-conf.\ wrong denotes examples with an incorrect final prediction
and confidence $\geq 0.9$.
}
\label{tab:smoothing}
\end{table}

Smoothing substantially improves calibration: ECE decreases from
0.206 to 0.147, Brier score from 0.225 to 0.207, and the fraction of
high-confidence wrong predictions from 11.0\% to 7.0\%.
In contrast, EM and F1 remain essentially unchanged, while AUROC
decreases only slightly from 0.795 to 0.790.
These results indicate that sparse CPT estimates primarily affect
calibration rather than the confidence-ranking signal.

\subsection{Statistical Significance Analysis}
\label{app:significance}

We assess the comparative claims in Sections~\ref{sec:stage2-main}--\ref{sec:stage2-boundary}
using paired bootstrap tests with 10,000 resamples, applying the same resampled
example indices to both methods in each comparison. Table~\ref{tab:significance}
summarizes the main tests.

\begin{table}[t]
\centering
\small
\resizebox{\linewidth}{!}{
\begin{tabular}{l l r r}
\toprule
Comparison & Metric & Difference & $p$-value \\
\midrule
Decomp-SC vs.\ CoT-SC
    & AUROC & $+0.004$ & $0.79$ \\
Graph-CoT-SC vs.\ CoT-SC
    & AUROC & $-0.002$ & $0.82$ \\
Sampled BN vs.\ CoT-SC
    & AUROC & $+0.029$ & $0.017$ \\
Sampled BN vs.\ CoT-SC
    & F1 & $+0.011$ & $0.16$ \\
\midrule
Sampled BN vs.\ CoT-SC, 4-hop
    & F1 & $-7.7$ pp & $<0.001$ \\
Sampled BN vs.\ CoT-SC, complex\_tree
    & F1 & $-12.3$ pp & $<0.001$ \\
Generated vs.\ gold edges, Sampled BN
    & F1 & $-5.8$ pp & $0.001$ \\
\bottomrule
\end{tabular}}
\caption{
Paired bootstrap significance tests for the main comparative claims in
Stage~2. Differences are computed as the first condition minus the second.
}
\label{tab:significance}
\end{table}

The AUROC differences between the graph-as-prompt methods and CoT-SC are not
significant, supporting the conclusion that graph structure provides little
ranking benefit when used only as prompt context. In contrast, Sampled BN
significantly improves AUROC over CoT-SC on the main Llama-3.1-70B setting,
while their overall F1 difference is not significant. The F1 degradation of
Sampled BN on 4-hop and \texttt{complex\_tree} examples as well as under
generated edges is significant. The corresponding AUROC differences on
4-hop and \texttt{complex\_tree} are not statistically significant, indicating
that the degradation on these difficult structures is primarily in answer
quality rather than confidence ranking.

% ============================================================
\subsection{Stage 2 Full Stratified Results}
\label{app:stage2-stratified-full}
% ============================================================

Table~\ref{tab:stage2-structure-compact} (main text) shows the compact CoT-SC versus Sampled~BN comparison across hop count and topology. Tables~\ref{tab:stage2-hop} and \ref{tab:stage2-topology-calibration} extend this to all seven methods.

% Table: Stage 2 hop-stratified results (Fuzzy column removed).
% Bold marks per-hop per-column best (highest for EM/F1/AUROC).
\begin{table}[t]
\centering
\small
\setlength{\tabcolsep}{4pt}
\resizebox{\linewidth}{!}{%
\begin{tabular}{l r r r r r}
\toprule
\textbf{Method}
  & \textbf{Hop}
  & \textbf{$N$}
  & \textbf{Ans.\ EM}$\uparrow$
  & \textbf{Ans.\ F1}$\uparrow$
  & \textbf{AUROC}$\uparrow$ \\
\midrule
Direct-SC      & 2 & 1252 & 0.491 & 0.602 & 0.717 \\
CoT-Conf       & 2 & 1252 & 0.522 & 0.636 & 0.688 \\
CoT-SC         & 2 & 1252 & 0.611 & 0.717 & 0.753 \\
Decomp-SC      & 2 & 1252 & 0.601 & 0.713 & 0.763 \\
Graph-CoT-SC   & 2 & 1252 & \textbf{0.617} & 0.727 & 0.762 \\
vPGM-Style     & 2 & 1252 & 0.466 & 0.587 & 0.608 \\
Sampled BN     & 2 & 1251 & 0.608 & \textbf{0.742} & \textbf{0.795} \\
\midrule
Direct-SC      & 3 & 760  & 0.450 & 0.560 & 0.701 \\
CoT-Conf       & 3 & 760  & 0.476 & 0.584 & 0.635 \\
CoT-SC         & 3 & 760  & 0.547 & 0.673 & 0.770 \\
Decomp-SC      & 3 & 760  & 0.507 & 0.627 & 0.758 \\
Graph-CoT-SC   & 3 & 760  & 0.516 & 0.632 & 0.763 \\
vPGM-Style     & 3 & 760  & 0.420 & 0.528 & 0.597 \\
Sampled BN     & 3 & 760  & \textbf{0.592} & \textbf{0.706} & \textbf{0.816} \\
\midrule
Direct-SC      & 4 & 405  & 0.402 & 0.487 & 0.757 \\
CoT-Conf       & 4 & 405  & 0.398 & 0.477 & 0.727 \\
CoT-SC         & 4 & 405  & \textbf{0.504} & \textbf{0.577} & 0.780 \\
Decomp-SC      & 4 & 405  & 0.479 & 0.559 & \textbf{0.814} \\
Graph-CoT-SC   & 4 & 405  & 0.486 & 0.568 & 0.778 \\
vPGM-Style     & 4 & 405  & 0.393 & 0.480 & 0.624 \\
Sampled BN     & 4 & 405  & 0.393 & 0.499 & 0.752 \\
\bottomrule
\end{tabular}}
\caption{Hop-stratified performance on MuSiQue full development set (2,417 examples, Llama-3.1-70B, $K{=}10$).
  Bold marks per-hop per-column best. Sampled BN leads on 2-hop and 3-hop
  but drops to 0.393 EM on 4-hop, where CoT-SC retains 0.504 EM.}
\label{tab:stage2-hop}
\end{table}

% Table: Stage 2 topology-stratified results with EM-based calibration metrics.
% Source files:
% - results/stage2_sc_baselines/llama31_70b_v2_full_dev/offline_reparsed_summary.csv
% - results/stage2_sc_baselines/llama31_70b_v2_full_dev/vpgm_official_answer_metrics.csv
% - results/section5/sampled_bn_full_dev_20260512/sampled_bn_official_answer_metrics.csv
% Topologies: simple_tree (1820), v_structure (351), complex_tree (246).
% Sampled BN simple_tree n=1819 because one example has a zero-valid required CPT row.
% AUROC/ECE/Brier are computed against Answer EM correctness. Higher is better for EM/F1/AUROC; lower is better for ECE/Brier.
% Bold marks per-topology per-column best.
\begin{table*}[!h]
\centering
\small
\setlength{\tabcolsep}{3.5pt}
\begin{tabular}{l l r r r r r r}
\toprule
\textbf{Method}
  & \textbf{Topology}
  & \textbf{$N$}
  & \textbf{Ans.\ EM}$\uparrow$
  & \textbf{Ans.\ F1}$\uparrow$
  & \textbf{AUROC}$\uparrow$
  & \textbf{ECE}$\downarrow$
  & \textbf{Brier}$\downarrow$ \\
\midrule
Direct-SC       & simple\_tree    & 1820 & 0.491 & 0.599 & 0.718 & 0.305 & 0.306 \\
CoT-Conf        & simple\_tree    & 1820 & 0.518 & 0.628 & 0.671 & 0.257 & 0.289 \\
CoT-SC          & simple\_tree    & 1820 & 0.597 & 0.713 & 0.761 & \textbf{0.160} & \textbf{0.220} \\
Decomp-SC       & simple\_tree    & 1820 & 0.587 & 0.697 & 0.752 & 0.221 & 0.245 \\
Graph-CoT-SC    & simple\_tree    & 1820 & 0.599 & 0.711 & 0.757 & 0.224 & 0.242 \\
vPGM-Style      & simple\_tree    & 1820 & 0.465 & 0.577 & 0.609 & 0.233 & 0.298 \\
Sampled BN      & simple\_tree    & 1819 & \textbf{0.605} & \textbf{0.737} & \textbf{0.789} & 0.200 & 0.220 \\
\midrule
Direct-SC       & v\_structure & 351  & 0.316 & 0.431 & 0.693 & 0.419 & 0.390 \\
CoT-Conf        & v\_structure & 351  & 0.356 & 0.464 & 0.683 & 0.355 & 0.348 \\
CoT-SC          & v\_structure & 351  & 0.430 & 0.519 & 0.761 & \textbf{0.215} & 0.247 \\
Decomp-SC       & v\_structure & 351  & 0.359 & 0.489 & \textbf{0.819} & 0.322 & 0.274 \\
Graph-CoT-SC    & v\_structure & 351  & 0.379 & 0.490 & 0.763 & 0.316 & 0.297 \\
vPGM-Style      & v\_structure & 351  & 0.319 & 0.433 & 0.616 & 0.283 & 0.299 \\
Sampled BN      & v\_structure & 351  & \textbf{0.456} & \textbf{0.555} & 0.790 & 0.226 & \textbf{0.246} \\
\midrule
Direct-SC       & complex\_tree   & 246  & 0.459 & 0.537 & 0.746 & 0.335 & 0.311 \\
CoT-Conf        & complex\_tree   & 246  & 0.439 & 0.517 & 0.748 & 0.259 & 0.278 \\
CoT-SC          & complex\_tree   & 246  & \textbf{0.589} & \textbf{0.661} & \textbf{0.796} & \textbf{0.098} & \textbf{0.186} \\
Decomp-SC       & complex\_tree   & 246  & 0.545 & 0.613 & 0.784 & 0.214 & 0.226 \\
Graph-CoT-SC    & complex\_tree   & 246  & 0.549 & 0.632 & 0.763 & 0.228 & 0.245 \\
vPGM-Style      & complex\_tree   & 246  & 0.423 & 0.514 & 0.646 & 0.190 & 0.265 \\
Sampled BN      & complex\_tree   & 246  & 0.451 & 0.538 & 0.792 & 0.228 & 0.237 \\
\bottomrule
\end{tabular}
\caption{Topology-stratified Answer EM, F1, and EM-based calibration metrics on MuSiQue full dev. Bold marks per-topology per-column best. Sampled BN leads on \textit{simple\_tree} and \textit{v\_structure} for answer quality, while CoT-SC remains strongest on \textit{complex\_tree}.}
\label{tab:stage2-topology-calibration}
\end{table*}

% Table: Stage 2 cross-model comparison: Qwen3-32B (non-reasoning) vs Llama-3.1-70B.
% Qwen source: results/stage2_sc_baselines/qwen3_32b_nonreasoning_full_dev/
%              results/section5/sampled_bn_qwen3_32b_nonreasoning_full_dev_20260513/
% Llama source: results/stage2_sc_baselines/llama31_70b_v2_full_dev/
%               results/section5/sampled_bn_full_dev_20260512/
% Qwen n=2417 (Sampled BN n=2414, 3 zero-valid-CPT examples excluded).
% Llama n=2417 (Sampled BN n=2416, 1 excluded). K=10, temperature 0.7, top_p=1.0.
% CoT-Conf uses K=1, temperature 0.
% Llama EM/F1/AUROC/ECE values match the source artifact re-aggregation;
%   may differ from Table~\ref{tab:stage2-main} by <=0.005 due to minor rounding.
% Bold marks per-model per-column best.
\begin{table*}[htbp]
\centering
\small
\resizebox{0.8\textwidth}{!}{%
\begin{tabular}{l rrrr rrrr}
\toprule
& \multicolumn{4}{c}{\textbf{Qwen3-32B}}
& \multicolumn{4}{c}{\textbf{Llama-3.1-70B}} \\
\cmidrule(lr){2-5} \cmidrule(lr){6-9}
\textbf{Method}
  & \textbf{EM}$\uparrow$
  & \textbf{F1}$\uparrow$
  & \textbf{AUROC}$\uparrow$
  & \textbf{ECE}$\downarrow$
  & \textbf{EM}$\uparrow$
  & \textbf{F1}$\uparrow$
  & \textbf{AUROC}$\uparrow$
  & \textbf{ECE}$\downarrow$ \\
\midrule
Direct-SC
  & 0.425 & 0.540 & 0.678 & 0.429
  & 0.462 & 0.569 & 0.718 & 0.325 \\
CoT-SC
  & \textbf{0.537} & 0.654 & \textbf{0.776} & \textbf{0.176}
  & \textbf{0.572} & 0.680 & 0.766 & \textbf{0.161} \\
CoT-Conf
  & 0.386 & 0.507 & 0.672 & 0.432
  & 0.487 & 0.593 & 0.686 & 0.271 \\
Decomp-SC
  & 0.534 & 0.653 & 0.722 & 0.327
  & 0.550 & 0.659 & 0.770 & 0.234 \\
Graph-CoT-SC
  & 0.536 & \textbf{0.656} & 0.717 & 0.331
  & 0.562 & 0.671 & 0.764 & 0.238 \\
vPGM-Style
  & 0.379 & 0.499 & 0.636 & 0.330
  & 0.439 & 0.550 & 0.622 & 0.235 \\
Sampled BN
  & 0.509 & 0.638 & 0.756 & 0.311
  & 0.568 & \textbf{0.691} & \textbf{0.795} & 0.206 \\
\bottomrule
\end{tabular}%
}
\caption{Cross-model Stage~2 comparison on MuSiQue full dev ($K{=}10$, temperature~0.7; CoT-Conf uses one temperature-0 call). Bold marks per-model per-column best.}
\label{tab:stage2-cross-model}
\end{table*}

% ============================================================
\subsection{Stage 2 Cross-Model Results}
\label{app:stage2-cross-model}
% ============================================================

Table~\ref{tab:stage2-cross-model} compares Qwen3-32B (non-thinking) and Llama-3.1-70B under the same MuSiQue full-dev Stage~2 protocol.

\paragraph{Stronger graph generation does not predict stronger uncertainty estimation.}
Qwen3-32B substantially outperforms Llama-3.1-70B on Stage~1 structural metrics (Setting~A Edge F1: 0.985 vs.\ 0.899) but falls behind on Stage~2 answer quality across most methods: CoT-SC EM is 0.537 for Qwen versus 0.572 for Llama, and Sampled~BN F1 is 0.638 versus 0.691. The Stage~1 dissociation between graph-structure quality and answer-generation quality therefore extends to Stage~2: better dependency wiring does not translate into better final answers or confidence.

\paragraph{Verbalized confidence is consistently weakest; CoT-SC calibration holds across models.}
For both models, vPGM-Style and CoT-Conf produce the lowest AUROC and worst calibration. CoT-SC is well-calibrated for both (ECE 0.176 for Qwen, 0.161 for Llama), and the graph-as-prompt methods add limited ranking gains over CoT-SC at the cost of overconfidence. These patterns replicate across models.

\paragraph{The Sampled BN versus CoT-SC ordering reverses.}
For Llama-3.1-70B, Sampled~BN leads on AUROC (0.795 vs.\ CoT-SC 0.766) and F1 (0.691 vs.\ 0.680). For Qwen3-32B, the ranking reverses: CoT-SC leads on AUROC (0.776 vs.\ Sampled~BN 0.756) and F1 (0.654 vs.\ 0.638). The graph-as-computation advantage that motivates Sampled~BN in the main text does not replicate for Qwen, indicating that the relative benefit of explicit BN propagation is model-dependent. The 4-hop degradation of Sampled~BN is consistent across both models and matches the structural sensitivity finding in the main text.

% ============================================================
% Table: 2WikiMultiHopQA Stage 2 cross-model diagnostic.
% Source: 2Wiki Stage 2 Llama vs Qwen report generated 2026-05-19.
% n=498 for all rows. SC/vPGM-Style/Sampled BN use K=10, temperature 0.7.
% CoT-Conf uses K=1, temperature 0.
\begin{table*}[!t]
\centering
\small
\resizebox{0.8\linewidth}{!}{%
\begin{tabular}{l rrrr rrrr}
\toprule
& \multicolumn{4}{c}{\textbf{Llama-3.1-70B}}
& \multicolumn{4}{c}{\textbf{Qwen3-32B}} \\
\cmidrule(lr){2-5} \cmidrule(lr){6-9}
\textbf{Method}
  & \textbf{EM}$\uparrow$
  & \textbf{F1}$\uparrow$
  & \textbf{AUROC}$\uparrow$
  & \textbf{ECE}$\downarrow$
  & \textbf{EM}$\uparrow$
  & \textbf{F1}$\uparrow$
  & \textbf{AUROC}$\uparrow$
  & \textbf{ECE}$\downarrow$ \\
\midrule
Direct-SC
  & 0.647 & 0.698 & 0.668 & 0.244
  & 0.582 & 0.653 & 0.619 & 0.359 \\
CoT-SC
  & \textbf{0.843} & \textbf{0.894} & 0.715 & \textbf{0.076}
  & 0.821 & \textbf{0.883} & \textbf{0.723} & \textbf{0.118} \\
CoT-Conf
  & 0.801 & 0.854 & 0.721 & 0.134
  & 0.769 & 0.854 & 0.636 & 0.165 \\
Decomp-SC
  & 0.799 & 0.853 & 0.715 & 0.139
  & 0.803 & 0.859 & 0.627 & 0.172 \\
Graph-CoT-SC
  & 0.805 & 0.857 & \textbf{0.730} & 0.133
  & 0.797 & 0.851 & 0.621 & 0.183 \\
vPGM-Style
  & 0.663 & 0.731 & 0.714 & 0.214
  & 0.631 & 0.778 & 0.463 & 0.278 \\
Sampled BN
  & 0.823 & 0.875 & 0.706 & 0.131
  & \textbf{0.827} & \textbf{0.883} & 0.611 & 0.152 \\
\bottomrule
\end{tabular}%
}
\caption{2WikiMultiHopQA Stage~2 diagnostic on the same 498-example stratified sample used in Stage~1 evaluation. Bold marks per-model best values.}
\label{tab:twowiki-stage2-cross-model}
\end{table*}

\subsection{2Wiki Stage 2 Cross-Dataset Diagnostic}
\label{app:stage2-twowiki}
% ============================================================

The 2Wiki Stage~2 diagnostic tests whether the MuSiQue Stage~2 pattern persists under a different task distribution. We use the same 498-example stratified sample as the Stage~1 diagnostic subset, with Llama-3.1-70B and Qwen3-32B non-reasoning evaluated under the same answer-metric protocol.

\paragraph{Answer quality converges; confidence ranking diverges by model.}
CoT-SC is the strongest single-method row for both models by answer F1 (Llama 0.894, Qwen 0.883), and Sampled~BN ties Qwen CoT-SC at F1~0.883 while Llama Sampled~BN reaches 0.875. The two models are therefore close on answer quality across most methods. Confidence ranking is a different story: Llama leads on AUROC for every graph-conditioned method, with the largest gaps on Graph-CoT-SC (0.730 vs.\ 0.621) and vPGM-Style (0.714 vs.\ 0.463). On Qwen, CoT-SC is the only method that achieves AUROC above 0.70 on 2Wiki.

\paragraph{Verbalized confidence is the most consistent failure mode.}
The 2Wiki results preserve the verbalized-confidence limitation from MuSiQue. Qwen vPGM-Style reaches F1~0.778 but AUROC~0.463, sharply separating answer quality from confidence ranking. Qwen CoT-Conf parses cleanly and reaches F1~0.854, yet its AUROC (0.636) remains below CoT-SC (0.723). Repeated sampling is a more reliable confidence source than verbalized probability elicitation on both datasets.

\paragraph{Structure difficulty on 2Wiki is governed by question type, not graph complexity.}
On MuSiQue, complex-topology questions are harder for both answer quality and graph generation. On 2Wiki, the pattern reverses: complex\_tree examples correspond to bridge\_comparison questions (4-step parallel chains), which achieve near-ceiling answer quality under CoT-SC (Llama EM 0.976, Qwen EM 0.958) because the rigid question template is easy to follow. Simple\_tree examples map to compositional and inference types, which have lower answer quality (CoT-SC F1 0.732 and 0.856 for Llama). This confirms that structural difficulty is governed by question-type distribution and annotation conventions rather than graph topology alone, consistent with the Stage~1 finding.

\end{document}